\documentclass[pdflatex,sn-nature]{sn-jnl}% Style for submissions to Nature Portfolio journals
\usepackage{graphicx}%
\usepackage{multirow}%
\usepackage{amsmath,amssymb,amsfonts}%
\usepackage{amsthm}%
\usepackage{mathrsfs}%
\usepackage[title]{appendix}%
\usepackage{xcolor}%
\usepackage{textcomp}%
\usepackage{manyfoot}%
\usepackage{booktabs}%
\usepackage{algorithm}%
\usepackage{algorithmicx}%
\usepackage{algpseudocode}%
\usepackage{listings}%
\usepackage{pifont}
\newcommand{\xmark}{\ding{55}}

\usepackage{makecell}

\usepackage{xcolor,colortbl}
\usepackage{color, colortbl} % https://texblog.org/2011/04/19/highlight-table-rowscolumns-with-color/ 
\usepackage{enumitem}       % modify indent in enumerate

\usepackage{tcolorbox}% http://ctan.org/pkg/tcolorbox
\usepackage{multirow}
\usepackage{colortbl}
\usepackage[table]{xcolor}

\theoremstyle{thmstyleone}%
\theoremstyle{thmstyletwo}%

\theoremstyle{thmstylethree}%
\definecolor{revblue}{rgb}{0.00,0.25,0.75}
\definecolor{revteal}{rgb}{0.00,0.45,0.45}

\newcommand{\rev}[1]{#1}
\newcommand{\revT}[1]{#1}
\newcommand{\revDel}[1]{}

\newcommand{\Sref}[1]{\S\ref{#1}}

\definecolor{highlight}{rgb}{0.9,0.95,1.0}	
\definecolor{Gray}{gray}{0.96}

\definecolor{lightBlue}{rgb}{0.78, 0.85, 1.0}% Rule colour
\definecolor{lightOrange}{rgb}{0.88, 0.95, 1.0}% Rule colour
\definecolor{lightRed}{rgb}{1.0, 0.85, 0.85}% Rule colour

\newtcbox{\bluebox}{on line, box align=base, colback=lightBlue,colframe=white,size=fbox,arc=3pt, before upper=\strut, top=-2pt, bottom=-4pt, left=-2pt, right=-2pt, boxrule=0pt}

\newtcbox{\orangebox}{on line, box align=base, colback=lightOrange,colframe=white,size=fbox,arc=3pt, before upper=\strut, top=-2pt, bottom=-4pt, left=-2pt, right=-2pt, boxrule=0pt}

\newtcbox{\redbox}{on line, box align=base, colback=lightRed,colframe=white,size=fbox,arc=3pt, before upper=\strut, top=-2pt, bottom=-4pt, left=-2pt, right=-2pt, boxrule=0pt}

\begin{document}

\title[Article Title]{A Multi-Instance LLM Framework for Rating the Quality of Surgical Feedback}


% Rafal , Everett, Steven, Jasmine, Cherine, Attharva, Ujjwal, Peter, Anima, Andrew

\author[1]{\fnm{Rafal} \sur{Kocielnik}}\email{rafal.kocielnik@gmail.com}

\author[3]{\fnm{J. Everett} \sur{Knudsen}}\email{everettknudsen@gmail.com}
% USC so USC/Norris Comprehensive Cancer Center

\author[3]{\fnm{Steven} \sur{Y. Cen}}\email{steven.cen@med.usc.edu}

\author[2]{\fnm{Jasmine} \sur{Lin}}\email{Jasmine.Lin@cshs.org}

\author[2]{\fnm{Cherine H.} \sur{Yang}}\email{cherine.yang@cshs.org}

\author[2]{\fnm{Atharva} \sur{Deo}}\email{atharva.deo@cshs.org}

\author[2]{\fnm{Ujjwal} \sur{Pasupulety}}\email{ujjwalpasupulety@gmail.com}

\author[2]{\fnm{Peter} \sur{Wager}}\email{peter.wager@cshs.org}

\author[1]{\fnm{Anima} \sur{Anandkumar}}\email{anima@caltech.edu}

\author*[1]{\fnm{Andrew J.} \sur{Hung}}\email{ajhung@gmail.com}

\affil[1]{\orgdiv{Computing + Mathematical Sciences}, \orgname{California Institute of Technology}, \orgaddress{\street{1200 E. California Blvd}, 
\city{Pasadena}, \postcode{91125}, 
\state{CA}, \country{USA}}}

\affil[2]{\orgdiv{Department of Urology}, \orgname{Cedars-Sinai}, \orgaddress{\street{8700 Beverly Blvd}, \city{Los Angeles}, \postcode{90048}, \state{CA}, \country{USA}}}

\affil[3]{\orgdiv{Keck School of Medicine}, \orgname{University of Southern California}, \orgaddress{\street{1500 San Pablo Street}, \city{Los Angeles}, \postcode{90033}, \state{CA}, \country{USA}}}


\abstract{Verbal feedback delivered by attending surgeons in the operating room plays a critical formative role in resident trainee skill acquisition. Yet, assessing the quality of trainer feedback and its effectiveness in influencing trainee behavior during live surgery remains a challenge. Prior studies assessed feedback content relying on extensive manual annotation by expert human raters and focused on developing broad taxonomies that overlook the qualitative aspects of feedback delivery such as clarity or urgency. Limited existing automated methods, including keyword analysis and topic modeling, also fail to capture these nuanced aspects. We introduce a two-stage LLM-based framework that discovers interpretable feedback quality criteria grounded in the context of surgical training. Our method uses \revT{multi-instance} prompting \rev{via repeated independent sampling of candidate criteria followed by consolidation,} and surgical domain knowledge injection to discover a small set of human interpretable scoring criteria (e.g., \emph{\revT{Encouragement}}, \emph{\revT{Urgency}}, \emph{\revT{Clarity}}). These criteria are then used to automatically score live surgical feedback via an LLM-as-a-judge approach. Evaluation on 4.2k trainer feedback instances demonstrates that our AI-discovered criteria outperform prior content-based frameworks in predicting feedback effectiveness, including observed trainee behavioral adjustments and trainer approval. This work advances scalable, human-aligned assessment of communication quality in the operating room and provides a foundation for improving surgical teaching practices.
}




\maketitle

\section{Introduction}\label{sec1}

% Importance & Problem
Formative verbal feedback to surgical trainees in the operating room (OR) plays a critical role in enhancing surgical education and outcomes \cite{agha2015role}. High-quality feedback during surgical training is associated with improved intraoperative performance \cite{bonrath2015comprehensive}, faster acquisition of technical skills \cite{ma2022tailored}, and greater trainee autonomy \cite{haglund2021surgical}. Feedback in the OR is typically triggered by a trainer’s observation of trainee behavior and is intended to shape future actions or decision-making. \revDel{The effectiveness of a feedback utterance lies in its ability to make a trainee adjust their behavior or verbally acknowledge the feedback in a manner that elicits trainer approval.} \rev{Within the broader context of surgical learning, we focus on observable trainee responses and trainer reactions as indicators of immediate feedback uptake that can be systematically assessed at scale. Specifically, we examine whether feedback is followed by trainee behavioral adjustment or verbal acknowledgment, and whether trainers subsequently express approval or disapproval.} Understanding how feedback is delivered—its clarity, urgency, timeliness, actionability, and emotional tone—is essential to improving its effectiveness in surgical training. Yet, systematically quantifying these aspects in live settings remains an open challenge.

\begin{figure}[t!]
    \centering
    \includegraphics[width=\linewidth]{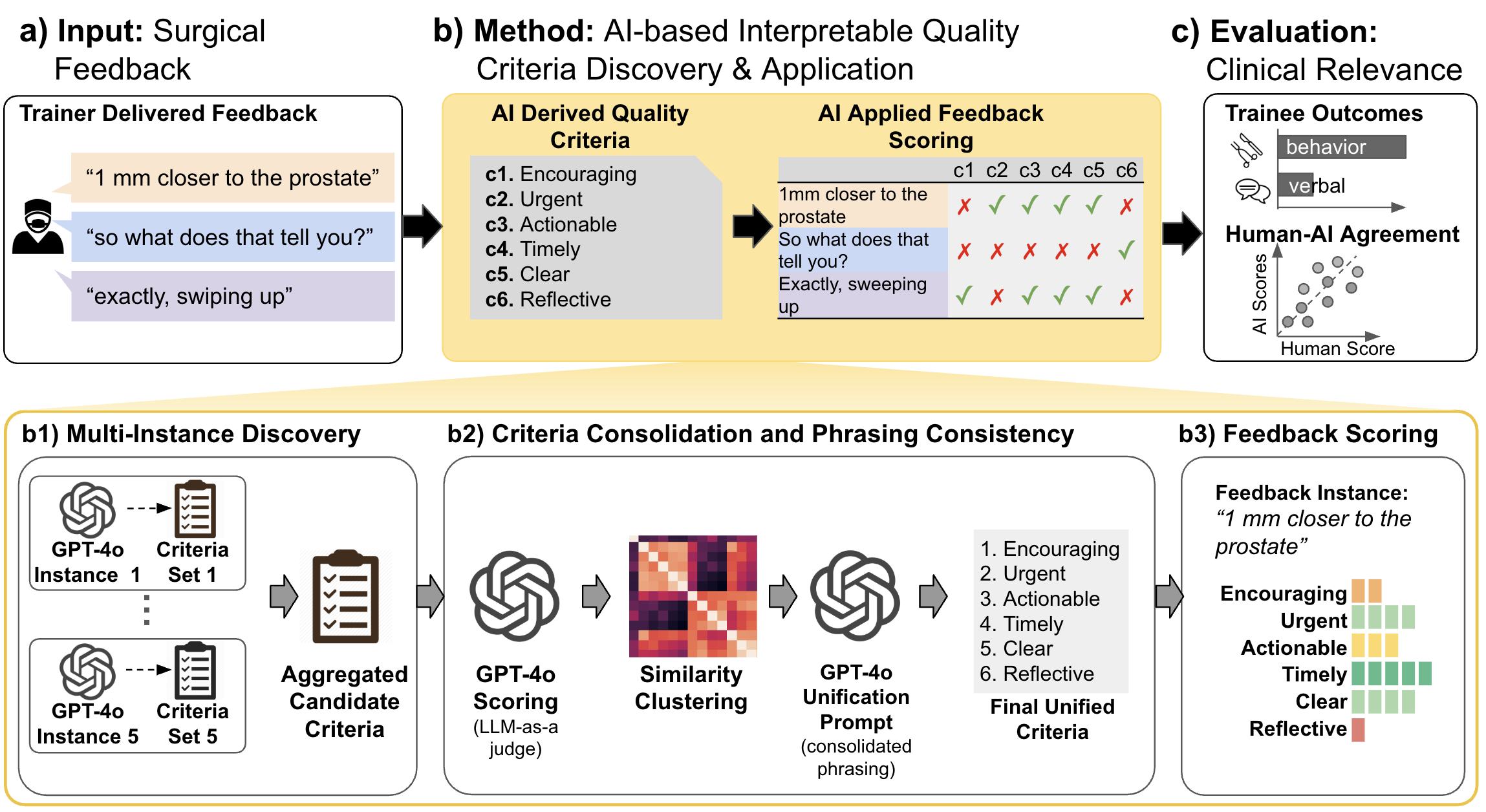}
    \caption{Overview of our AI-based framework for interpretable discovery and evaluation of surgical feedback quality. 
    (a) Trainer-delivered feedback during live procedures is used as input. 
    (b) The method identifies and scores core feedback quality criteria. First, multiple GPT-4o \revT{instances} generate candidate criteria sets (b1), which are then consolidated (b2) using LLM scoring, similarity clustering, and prompt-based unification into six core criteria: \emph{\revT{Encouragement}}, \emph{\revT{Urgency}}, \emph{\revT{Actionability}}, \emph{\revT{Timeliness}}, \emph{\revT{Clarity}}, and \emph{\revT{Reflection}}. Each feedback instance is scored along these dimensions (b3) using LLM-as-a-judge approach.
    (c) Scored feedback is evaluated for clinical relevance through its association with observed trainee outcomes (verbal and behavioral) and agreement with human annotations.}
    \label{fig:task_overview}
\end{figure}

% Prior Work
Previous research on surgical feedback has primarily focused on categorizing what instructors communicate during procedures. This includes typologies of trainer command types (e.g., guiding, questioning, chastising) \citep{hauge2001reliability}, thematic content such as anatomy or instrument handling \citep{blom2007analysis}, discourse structures and planning strategies \citep{d2020evaluating}, and broad feedback categories (e.g., procedural, technical, praise, criticism) \citep{wong2023development}. These frameworks have advanced understanding of instructional goals and content in surgical education. However, significantly less attention has been given to how feedback is delivered and interpreted. Systematic analysis is also difficult due to the combined need for specialized domain knowledge and labor-intensive annotation processes. Some automation methods have been proposed, including keyword-frequency techniques (e.g., LIWC \citep{ramprasad2024language}) and topic modeling with language model embeddings \citep{kocielnik2024human}\rev{, as well as multi-stage pipelines that detect feedback in operating-room audio and assign it to established content categories \citep{nasriddinov2025automating}. These systems automate \emph{what} was said and which category it falls into, but leave the delivery dimensions unquantified}. LIWC lacks sensitivity to clinical language, relies on predefined keyword dictionaries, and cannot capture delivery-focused attributes such as clarity, urgency, or instructional tone. Topic modeling approaches like BERTopic \citep{grootendorst2022bertopic} cluster semantically related content (e.g., feedback on procedures such as \textit{``sweeping and cutting''} or \textit{``needle positioning''}) but similarly fail to distinguish meaningful differences in how feedback is delivered even in identical instructional contexts. For example, a directive like \textit{``Move the needle to the left now''} carries a different instructional tone and urgency compared to \textit{``Let’s try adjusting the needle slightly to the left''}, despite both referring to the same task (\textit{“instrument handling”}). These delivery aspects are critical for understanding feedback effectiveness from the trainee’s perspective.

%%% Our novel approach
To address these challenges, we introduce an LLM-based framework for analyzing the delivery quality of surgical feedback (Figure~\ref{fig:task_overview}). The framework takes as input (a) samples of raw transcripts of trainer-delivered verbal feedback during live surgical cases together with clinical domain knowledge (i.e., clinically validated definitions of feedback and effectiveness outcome criteria from \cite{wong2023development}). This input is passed to a large language model to discover and operationalize core feedback quality criteria (b). In the \textbf{discovery phase (b1)}, we use \revT{multi-instance} prompting with separate GPT-4o instances \rev{(independent samples from the same model, aggregated at the consolidation stage)}, where \revT{instances} are seeded with both formal definitions and a representative subset of feedback examples to independently propose candidate evaluation criteria. In the \textbf{criteria consolidation phase (b2)}, these candidate criteria are then applied to feedback, clustered based on scoring similarity and unified into one phrasing through an additional GPT-4o consolidation prompt, producing a small set of stable, human-interpretable, and domain-grounded dimensions: \emph{\revT{Encouragement}}, \emph{\revT{Urgency}}, \emph{\revT{Actionability}}, \emph{\revT{Timeliness}}, \emph{\revT{Clarity}}, and \emph{\revT{Reflection}}. In the \textbf{scoring phase (b3)}, the final criteria are applied to individual feedback instances using an LLM-as-a-judge setup, enabling automated scoring with human-interpretable rubric. We assess the clinical relevance of the criteria (c) by assessing their ability to predict behavioral outcomes of trainer and trainee, in comparison to prior work; evaluating their alignment with human reasoning; and uncovering how different feedback quality properties lead to particular behavioral outcomes.
This step validates the practical impact of feedback quality dimensions in real-world settings and their interpretability. Our framework enables fine-grained, scalable, and interpretable evaluation of feedback delivery, \rev{allowing automated quality scoring after initial human calibration and validation without requiring manual quality annotation of every feedback instance. Our methodological contribution is an integrated pipeline that uses pretrained LLMs and clinical domain knowledge to discover, consolidate, and apply human-interpretable feedback quality criteria. Downstream supervised models evaluate the predictive utility of the resulting scores for observed trainee and trainer responses.} \revDel{Our approach departs from prior work in several key technical ways. Rather than relying on predefined taxonomies or unsupervised clustering methods such as BERTopic, we adopt a two-stage, LLM-guided strategy and combine it with prior clinical knowledge.}

%%% Findings
Our framework uncovers six interpretable feedback quality criteria—such as clarity, actionability, and urgency—that effectively predict trainee \revT{behavioral adjustment}. The six quality criteria by themselves produce high AUC scores ranging from 0.71 to 0.75 for predicting trainee reaction to feedback (Table \ref{tab:predicting_outcomes}). Competitive analysis against existing feedback categorization frameworks from prior work, revealed that our criteria consistently improve prediction of trainee behavior by 9–12\% and trainer reaction by 3-11\%. In combination with content categories from prior work, our quality criteria reach AUCs ranging from 0.74 to 0.78 for trainee outcome prediction. Human scoring of feedback using the discovered criteria aligns substantially with LLM-applied scores (weighted $\kappa$ = 0.60–0.70) for 5 of the 6 criteria, underscoring their interpretability and practical usability. On the methodological side, our results highlight the value of \revT{multi-instance} prompting and consolidation steps for discovery of novel human-interpretable scoring criteria. This extends the prior approaches, such as LLM-as-a-judge relying on fixed rubric representing a priori provided criteria. 

% Impact statement
By enabling automated, interpretable assessment of feedback delivery grounded in real-world behavioral outcomes, our framework offers a practical tool for improving intraoperative teaching effectiveness and supporting trainer development. Beyond individual evaluations, this approach lays the groundwork for scalable integration into surgical education pipelines and clinical quality assurance systems, advancing the broader goal of optimizing communication-driven learning in high-stakes healthcare environments.

%%%%%%%%%%%%%%%%%
%%%% RESULTS %%%%
%%%%%%%%%%%%%%%%%
\section{Results: Clinical Interpretation and Validation}
\label{sec:results}

Our process led to the discovery of 6 interpretable feedback quality criteria rated on a 5-point Behavioral Anchored Rating Scales (BARS). Using LLM-as-judge approach \citep{zheng2023judging}, where an LLM is asked to rate feedback using provided criteria and scales, we applied GPT-4o to rate 4210 lines of live surgical feedback collected in prior work \citep{wong2023development}. We subsequently evaluated these quality criteria ratings for their ability to predict clinical effectiveness of feedback in affecting trainee behavior and leading to subsequent approval from a trainer. In this evaluation, we compared our AI discovered quality criteria to automated topic modeling approach from recent work \citep{kocielnik2024human} and fully manually annotated human expert proposed categories \citep{wong2023development}. \rev{We then} evaluated the ability of human raters with domain knowledge to use these AI-discovered quality criteria to rate the feedback instances consistently. \rev{We next} analyzed the statistical associations of the individual quality criteria with trainee behavioral adjustment and trainee verbal acknowledgment to understand how each quality criterion affects outcomes. \rev{We close with robustness checks on the criteria themselves: their mutual independence, the sensitivity of the results to how cross-validation folds are formed, and whether the rubric can be applied by models other than GPT-4o.}
Further details can be found in the \emph{Methods} section.

%%% Feedback Effectiveness Table %%
\begin{table}[t!]
\centering
\begin{tabular}{
>{\raggedright\arraybackslash}p{4.7cm} 
p{1.5cm} 
p{1.5cm} 
p{1.5cm}
p{1.6cm}
}
\toprule
\textbf{Feedback Criteria} & 
\multicolumn{2}{c}{\shortstack{\textbf{Trainee Reaction} \\ \textbf{to Feedback}}} & 
\multicolumn{2}{c}{\shortstack{\textbf{Final Trainer Reaction} \\ \textbf{to Trainee Reaction}}} \\
\cmidrule(lr){2-3} \cmidrule(lr){4-5}
& \textbf{\revT{Behavioral Adjustment}} & \textbf{\revT{Verbal Acknowledgment}} & \textbf{Approval} & \textbf{Disapproval} \\
\midrule
AI-derived Quality Scores & 0.75$_{\pm0.01}$ & 0.71$_{\pm0.02}$ & 0.66$_{\pm0.02}$ & 0.60$_{\pm0.06}$\\
\midrule
Prior Topic Modeling~\cite{kocielnik2024human} & 0.69$_{\pm0.02}$ & 0.66$_{\pm0.02}$ & 0.67$_{\pm0.02}$ & 0.57$_{\pm0.06}$\\
+ AI Quality Scores & 0.77$^{*}_{\pm0.01}$ & 0.73$^{*}_{\pm0.02}$ & 0.69$^{\rev{\dagger}}_{\pm0.02}$ & 0.63$^{*}_{\pm0.06}$ \\
\midrule
Prior Human-defined Categories~\cite{wong2023development} & 0.70$_{\pm0.02}$ & 0.68$_{\pm0.02}$ & 0.63$_{\pm0.02}$ & 0.59$_{\pm0.06}$\\
+ AI Quality Scores & 0.78$^{*}_{\pm0.01}$ & 0.74$^{*}_{\pm0.01}$ & 0.69$^{*}_{\pm0.02}$ & 0.62$_{\pm0.06}$ \\
\bottomrule
\end{tabular}
\vspace{0.5em}
\caption{\textbf{Predicting trainee and trainer behavioral outcomes in reaction to feedback.} Performance of LLM-driven interpretable scoring criteria discovery compared to prior approaches. We report mean AUROC ± 95\% CI for predicting four behavioral outcomes from Wong et al.~\cite{wong2023development}: Behavior Adjustment, Verbal Acknowledgment, Trainer Approval, and Trainer Disapproval. AI-derived quality criteria are evaluated alone and in combination with prior topic modeling~\cite{kocielnik2024human} and human-defined feedback categories~\cite{wong2023development}. 
Statistical significance ($^*$\,$p < 0.05$) indicates that adding AI-derived quality scores led to a significant improvement, assessed via DeLong’s test, \rev{with the entire 95\% CI of the AUROC difference lying above 0. $\dagger$ denotes a marginal improvement for which the lower confidence bound lies above 0 by less than 0.001.}}
\label{tab:predicting_outcomes}
\end{table}

\subsection{\rev{Prediction of Trainee and Trainer Responses}}
We evaluated the predictive performance of six LLM-derived feedback quality ratings across four behavioral outcomes using fivefold stratified cross-validation with Random Forest classifiers (Table \ref{tab:predicting_outcomes}). 

% Quality - Table 1
Models using only the six quality ratings achieved strong performance, including AUROC=0.75, 95\% CI: [0.74, 0.77] for \emph{Trainee \revT{Behavioral Adjustment}}, 0.71 [0.69, 0.72] for \emph{Trainee \revT{Verbal Acknowledgment}}, 0.66 [0.64, 0.68] for \emph{Trainer Approval}, and 0.60 [0.54, 0.66] for \emph{Trainer Disapproval} of Trainee Reaction. Augmenting prior topic modeling categories~\cite{kocielnik2024human} with our AI-derived quality scores led to consistent improvements across all outcomes: AUROC=0.77 [0.76, 0.79] (+12\% gain), 0.73 [0.71, 0.74] (+9\%), 0.69 [0.67, 0.71] (+3\%), and 0.63 [0.58, 0.69] (+11\%), respectively. Similarly, augmenting prior manually annotated Human Categories~\cite{wong2023development} with our AI-derived quality ratings provided consistent gains in predictive performance, yielding AUROC=0.78 [0.77, 0.80] (+12\% gain over human proposed categories) for trainee behavior adjustment, 0.74 [0.73, 0.76] (+9\%) for verbal acknowledgment, 0.69 [0.67, 0.71] (+9\%) for trainer approval, and 0.62 [0.56, 0.68] (+5\%) for trainer disapproval.

% DeLong's comparisons
To assess the significance of these gains, we applied DeLong’s test for correlated AUROC curves (Table~\ref{tab:delong_comparative_analysis} in Methods). \rev{Relative to models using only Topic Modeling features, adding AI quality scores improved AUROC for \emph{Trainee Behavioral Adjustment} ($\Delta$=+0.080, 95\% CI: [0.066, 0.093]) and \emph{Trainee Verbal Acknowledgment} (+0.061 [0.048, 0.074]), with both intervals lying above zero. The corresponding gains were +0.065 [0.004, 0.127] for \emph{Trainer Disapproval}, with the interval above zero, and +0.017 [0.000, 0.035] for \emph{Trainer Approval}, whose lower bound lies above zero by less than 0.001, and which we therefore regard as marginal. Relative to Human Categories, gains were +0.082 [0.067, 0.096] for \emph{Trainee Behavioral Adjustment}, +0.064 [0.051, 0.078] for \emph{Trainee Verbal Acknowledgment}, and +0.058 [0.036, 0.079] for \emph{Trainer Approval}, all with intervals above zero; the gain for \emph{Trainer Disapproval} was +0.026 [$-$0.033, 0.085] and did not exclude zero. AI-derived quality dimensions therefore add predictive value for both trainee outcomes against both baselines, and for each trainer outcome against one of the two baselines. Across outcomes they complement both automated content-based and manual expert-coded feedback aspects.}

%%%%%%%%%%%% HUMAN-AI Alignment %%
\begin{table}[t!]
\centering
\small
\begin{tabular}{lcc@{\hskip 0.8cm}cc@{\hskip 0.8cm}cc}
\toprule
\textbf{Quality Criterion} 
& \multicolumn{2}{c}{\textbf{Human-Human}} 
& \multicolumn{2}{c}{\textbf{AI-AI}} 
& \multicolumn{2}{c}{\textbf{Human-AI}} \\
\cmidrule(lr){2-3} \cmidrule(lr){4-5} \cmidrule(lr){6-7}
& \textbf{K} & \textbf{95\% CI} & \textbf{K} & \textbf{95\% CI} & \textbf{K} & \textbf{95\% CI} \\
\midrule
Encouragement      & 0.79 & (0.46, 0.91) & 1.00 & (1.00, 1.00) & 0.72 & (0.42, 0.86) \\
Urgency            & 0.72 & (0.52, 0.85) & 0.98 & (0.94, 1.00) & 0.68 & (0.48, 0.82) \\
Actionability      & 0.76 & (0.55, 0.88) & 1.00 & (1.00, 1.00) & 0.79 & (0.63, 0.89) \\
Timeliness         & 0.44 & (0.24, 0.58) & 0.94 & (0.62, 1.00) & 0.54 & (0.32, 0.76) \\
Clarity            & 0.67 & (0.47, 0.82) & 0.92 & (0.79, 1.00) & 0.75 & (0.49, 0.92) \\
Reflection         & 0.71 & (0.42, 0.87) & 0.98 & (0.86, 1.00) & 0.74 & (0.41, 0.90) \\
\bottomrule
\end{tabular}
\vspace{0.5em}
\caption{\textbf{Agreement analysis across combinations of Human and AI raters.}} Quadratic Weighted Kappa (\textbf{K}) scores and 95\% confidence intervals (CIs) for quality scoring agreement across three rater configurations: two human raters (Human-Human), two  AI runs (AI-AI), and average human vs. AI scoring (Human-AI). Score interpretation thresholds: 0.01–0.20 (slight), 0.21–0.40 (fair), 0.41–0.60 (moderate), 0.61–0.80 (substantial), 0.81–1.00 (almost perfect agreement) \citep{landis1977measurement}.
\label{tab:kappa_agreement}
\end{table}

\subsection{Alignment of AI scoring with Human Annotations}
To evaluate alignment with human judgment, we took the AI-discovered quality rating definitions and asked two human raters with domain knowledge to apply them to 30 randomly selected feedback instances. Raters received a training session using a separate set of 30 examples. Further details can be found in the \emph{Methods} section.

We evaluated inter-rater reliability across three configurations using quadratically weighted Cohen’s kappa~\cite{McHugh2012}, which is appropriate for ordinal scales such as BARS. (Table~\ref{tab:kappa_agreement}): between two human raters (Human-Human), between two AI runs (AI-AI), and between the AI and the averaged scores of the human raters (Human-AI).

We observe substantial agreement among human raters across most dimensions (e.g., Encouragement: $K=0.79$, 95\% CI: [0.46, 0.91]; Actionability: $K=0.76$, CI: [0.55, 0.88]; Urgency: $K=0.72$, CI: [0.52, 0.85]). Agreement was lower for more subjective and contextual dimensions such as Timeliness ($K=0.44$, CI: [0.24, 0.58]), suggesting inherent difficulty in consistently judging temporal aspects of feedback.

AI-generated ratings showed near-perfect internal consistency across repeated runs (AI-AI: $K=0.92$–1.00), indicating deterministic and stable behavior. We note that these have been collected under the temperature setting of 0.0 to encourage deterministic behavior. Further details of AI setup using GPT-4o are provided in the \emph{Methods} section. Importantly, Human-AI agreement was also substantial across most criteria (e.g., Actionability: $K=0.79$, CI: [0.63, 0.89]; Clarity: $K=0.75$, CI: [0.49, 0.92]), approaching inter-human agreement levels. This suggests that the AI model is not only consistent in its ratings but also well-aligned with expert human judgment, particularly on dimensions that are less subjective or more structurally grounded in language.

\begin{figure}[t!]
    \centering
    \includegraphics[width=\linewidth]{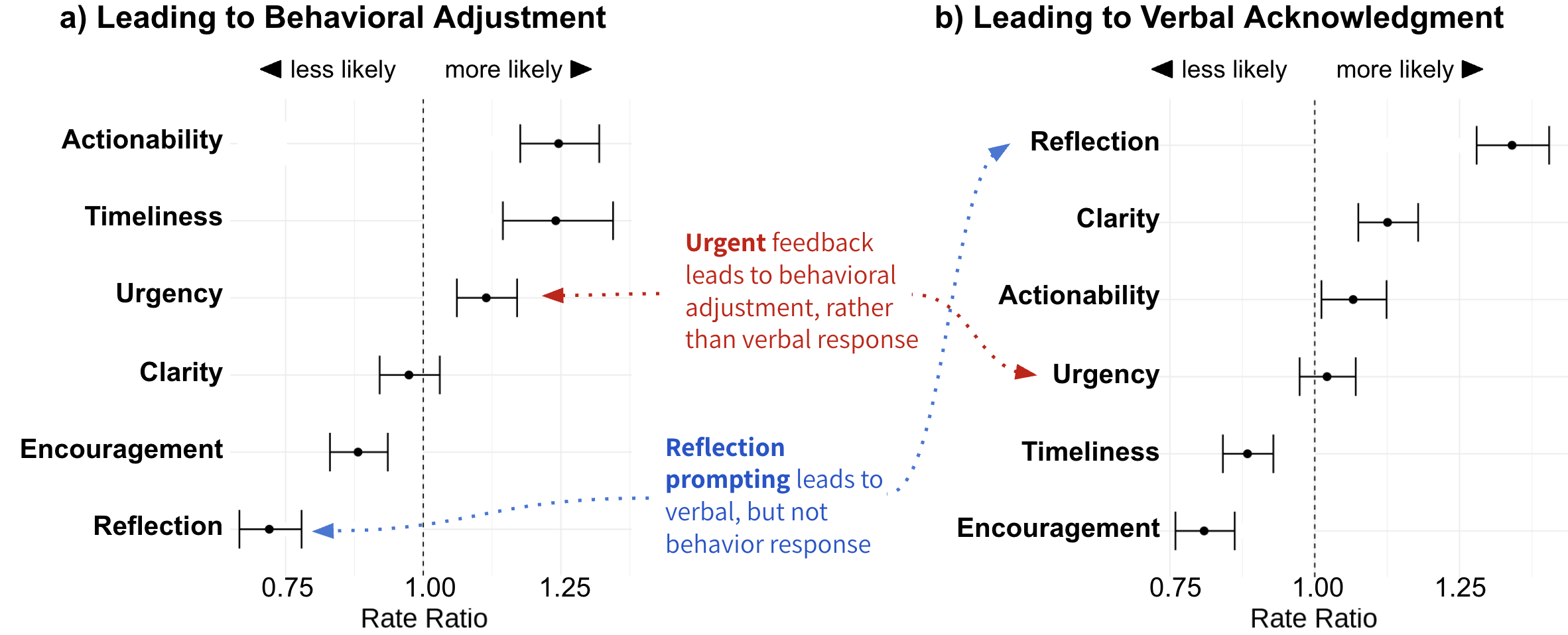}
    \caption{\textbf{Associations between discovered feedback quality criteria and trainee behavioral outcomes.} Rate ratios and confidence intervals are shown for each of the six LLM-discovered feedback quality dimensions in relation to (a) \emph{Trainee Behavioral Adjustment} and (b) \emph{Trainee Verbal Acknowledgment} outcomes. Timely feedback was most predictive of \revT{behavioral adjustment}, while reflective and encouraging feedback were more strongly associated with verbal acknowledgment but not behavior change. These findings highlight how distinct delivery qualities of surgical feedback differentially influence trainee responses.}
    \label{fig:quality_associations}
\end{figure}

\subsection{Criteria Association with Feedback Effectiveness}

To understand the real-world impact of our discovered criteria, we next examine how each quality dimension relates to trainee behavioral adjustments and verbal acknowledgments following feedback. \rev{The rate ratios below summarize each dimension with a single multiplicative effect; the corresponding marginal probabilities at every point of the 1--5 rating scale are shown in Appendix~\ref{apx:marginal_effects}, Figures~\ref{fig:marginal_effects} and~\ref{fig:marginal_effects_verbal}.}

\emph{Trainee Behavioral Adjustment} was significantly associated with several feedback quality dimensions (Fig.~\ref{fig:quality_associations}a). Feedback rated highly on \emph{\revT{Actionability}} (Rate Ratio [RR] = 1.22, 95\% CI: [1.18, 1.32]), \emph{\revT{Timeliness}} (RR = 1.24, CI: [1.14, 1.34]), and \emph{\revT{Urgency}} (RR = 1.11, CI: [1.06, 1.17]) was associated with higher rates of observed \revT{trainee behavioral adjustment}. These dimensions reflect feedback that is specific, timely, and emphasizes the need for immediate action—elements that are directly conducive to real-time correction of performance. In contrast, \emph{\revT{Encouragement}} feedback (RR = 0.88, CI: [0.83, 0.94]) and \emph{\revT{Reflection}} feedback (RR = 0.72, CI: [0.67, 0.78]) were associated with reduced behavioral adjustment. This may be due to the nature of encouraging feedback, which often affirms correct behavior without requiring further adjustment, and reflective feedback, which aims to stimulate longer-term insight rather than immediate correction. \emph{Clarity} did not show a statistically significant effect (RR = 0.97, CI: [0.92, 1.03]).

\emph{Trainee Verbal Acknowledgment} showed a distinct pattern of associations with feedback quality dimensions (Fig.~\ref{fig:quality_associations}b). Feedback rated highly on \emph{\revT{Reflection}} (Rate Ratio [RR] = 1.34, 95\% CI: [1.28, 1.40]) and \emph{\revT{Clarity}} (RR = 1.13, CI: [1.08, 1.18]) was associated with higher rates of verbal acknowledgment. These findings suggest that verbal reactions are more likely when trainees are prompted to think or when feedback is easily understood. \rev{High} \emph{\revT{Actionability}} feedback showed a smaller but significant positive association (RR = 1.07, CI: [1.01, 1.12]). In contrast, \rev{high} \emph{\revT{Encouragement}} feedback (RR = 0.81, CI: [0.76, 0.86]) and feedback \rev{rated highly on} \emph{\revT{Timeliness}} (RR = 0.88, CI: [0.84, 0.93]) were associated with reduced acknowledgment. Again, encouraging feedback may act as affirmation, often concluding an interaction rather than prompting a response, while timely feedback may be delivered in fast-paced moments when verbal acknowledgment is less feasible. \emph{Urgency} was not significantly associated with this outcome (RR = 1.02, CI: [0.97, 1.07]).

These results confirm distinct patterns of feedback effectiveness across outcomes. While \emph{\revT{Actionability}}, \emph{\revT{Timeliness}}, and \emph{\revT{Urgency}} \rev{of feedback} increased behavioral response rates, \emph{\revT{Reflection}} and \emph{\revT{Clarity}} \rev{of feedback} were stronger predictors of verbal acknowledgment. \emph{\revT{Encouragement}} \rev{present in feedback} consistently decreased the likelihood of both response types. Interestingly, \emph{\revT{Timeliness}} \rev{of feedback} had opposite effects—positively associated with behavior but negatively with acknowledgment—suggesting that different delivery styles selectively influence trainee behavior.

%% Word Clouds around Criteria
\begin{figure}[t!]
    \centering
    \includegraphics[width=\linewidth]{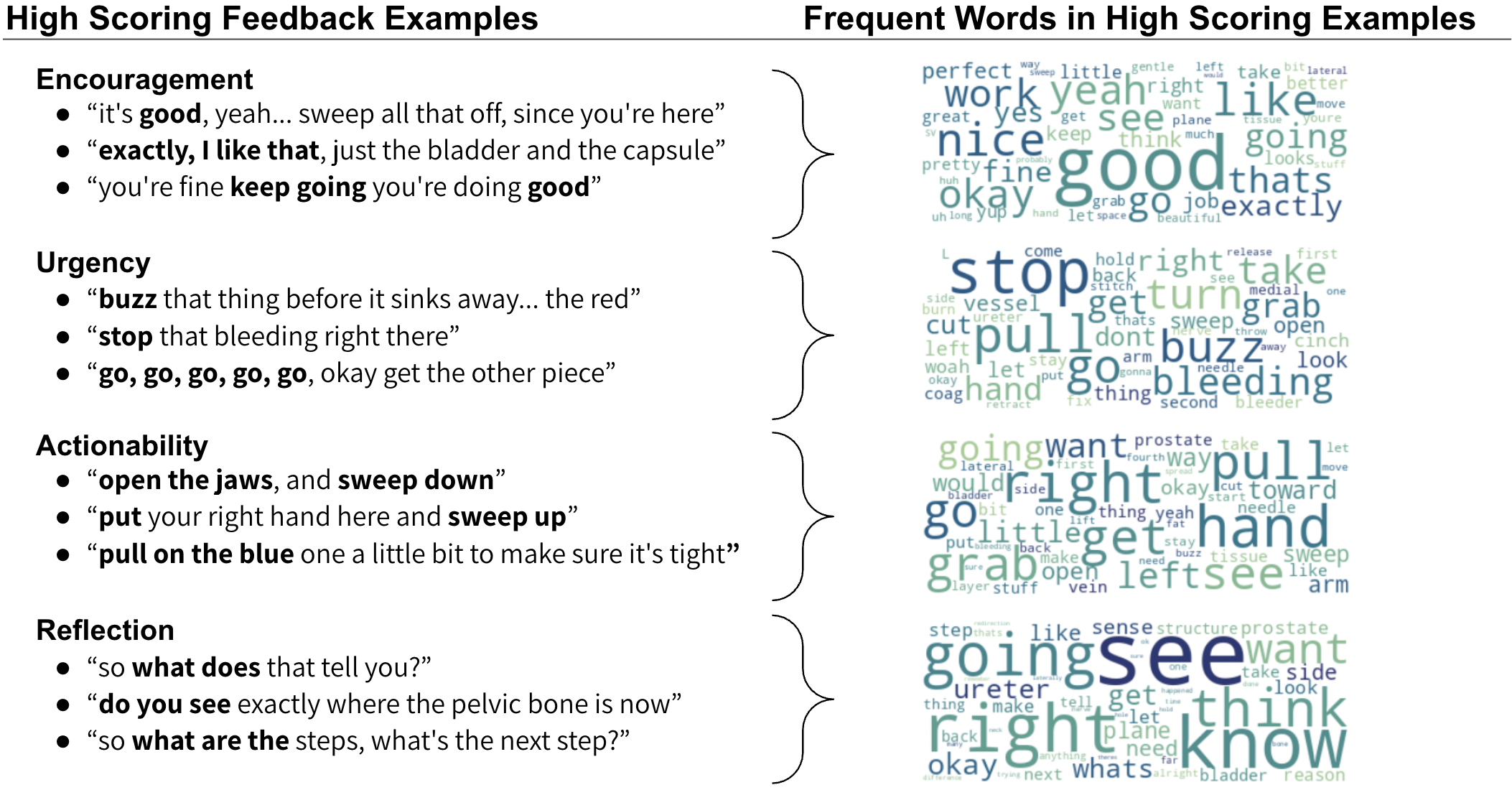}
    \caption{Illustrative examples and linguistic patterns associated with high-scoring feedback. \textbf{Left:} Representative trainer feedback excerpts scoring highly on four of the discovered quality dimensions—\emph{Encouragement}, \emph{Urgency}, \emph{Actionability}, and \emph{Reflection}—demonstrate the linguistic structure and tone aligned with each dimension. \textbf{Right:} Word clouds visualize the most frequent terms in high-scoring instances across all criteria, highlighting key lexical patterns (Encouragement : ``good'', Urgency: ``stop'', Actionability: ``pull'', Reflection : ``see'') associated with effective feedback delivery in surgical training.}
    \label{fig:quality_wordclouds}
\end{figure}

%%%%%%%%%%%% ROBUSTNESS CHECKS (R2) -- kept last: supporting, not headline, results %%%%%%%%%%%%
\subsection{\rev{Robustness Checks on the Discovered Criteria}}
\label{sec:robustness_checks}

\rev{We report three checks on the criteria themselves: whether the six dimensions are statistically distinct, whether performance depends on how cross-validation folds are formed, and whether the rubric can be applied by scoring models other than GPT-4o. Full results are in Appendix~\ref{apx:correlation_structure}.}

\rev{\emph{The six dimensions are only weakly intercorrelated.} Across the 15 unique pairs of GPT-4o criterion scores, the mean absolute Spearman correlation was 0.26, and the strongest pair---\emph{Urgency} and \emph{Actionability}---reached 0.63 (Table~\ref{tab:spearman_criteria}). Maximum variance inflation factors were 2.25 for GPT-4o, 3.29 for Mistral, and 3.95 for Qwen (Table~\ref{tab:vif_grouping}), below conventional thresholds for problematic multicollinearity. The criteria therefore capture largely distinct aspects of feedback delivery rather than repeated measurements of a single underlying construct.}

\rev{\emph{Results do not depend on how folds are formed.} Comparing outcome-stratified with case-grouped cross-validation, in which all utterances from a given surgical case are kept together, AUROC differences ranged from $-0.012$ to $+0.002$ across models and outcomes (Table~\ref{tab:vif_grouping}).}

\rev{\emph{Open-weight models match GPT-4o's predictive performance, but agree with it only moderately on individual scores.} Scoring the same corpus with the finalized rubric, \emph{Behavioral Adjustment} AUROCs were 0.741 for GPT-4o, 0.727 for Mistral Small 3.2, and 0.731 for Qwen3-Next; the corresponding \emph{Verbal Acknowledgment} AUROCs were 0.692, 0.699, and 0.675 (Table~\ref{tab:cross_model}A). Item-level agreement was lower: mean $\kappa$ with a GPT-4o rerun was 0.528 (95\% CI: [0.514, 0.541]) for Mistral and 0.600 [0.587, 0.612] for Qwen, both moderate on conventional thresholds~\citep{landis1977measurement}. Scores fell within one point of GPT-4o in 89.1\% [88.0\%, 90.1\%] and 90.5\% [89.7\%, 91.2\%] of comparisons, and agreement ranged from substantial for \emph{Urgency} and \emph{Actionability} to fair for \emph{Timeliness} in both models (Table~\ref{tab:cross_model}B).}

\section{Discussion}
\label{sec:discussion}
The quality of verbal feedback in surgical training is essential for guiding real-time performance and promoting long-term skill development among surgical trainees \cite{agha2015role, bonrath2015comprehensive, ma2022tailored}. Despite its importance, efforts to understand and evaluate feedback in the operating room (OR) have focused primarily on content categories—such as communication type \cite{blom2007analysis}, teaching behavior \cite{hauge2001reliability}, or content themes \cite{wong2023development}, with limited focus on its quality from a trainee's perspective. These approaches also often rely on labor-intensive manual annotation \citep{wong2023development}, limited human-derived recognition of important delivery patterns \citep{quesada2014examination}, and lack robust validation linking feedback to behavioral outcomes \citep{wong2023development, d2020evaluating, blom2007analysis, hauge2001reliability}.

Assessing the \emph{quality} of surgical feedback—how it is delivered rather than just what is said—is critical for improving training outcomes. However, existing frameworks often overlook delivery dimensions like urgency, clarity, and timeliness, which are crucial for feedback effectiveness. Automated methods like topic modeling or keyword analysis fail to capture these nuances, and human-driven approaches, aside from largely omitting these aspects, are also not scalable. Our work addresses these gaps by introducing a large language model (LLM)-based framework that discovers and scores interpretable feedback quality dimensions, enabling scalable, clinically grounded evaluation aligned with trainee behavior.

Our approach surfaced a set of interpretable, behaviorally grounded feedback quality criteria that were both emergent and predictive across surgical training interactions. These criteria—\emph{Encouragement}, \emph{Urgency}, \emph{Actionability}, \emph{Clarity}, \emph{Timeliness}, and \emph{Reflection}—capture distinct pragmatic and pedagogical dimensions of trainer communication. Each dimension reflects a unique function: Encouragement denotes feedback that is supportive and provides positive reinforcement to boost confidence; Urgency reflects the communication that immediate action or attention is required; Actionability refers to clear, specific actions or steps the trainee can implement; Timeliness captures whether feedback is provided during or promptly after the trainee’s action or decision; Clarity assesses whether the message is straightforward, unambiguous, and easily understood; and Reflection involves prompting the trainee to self-assess or reflect on their performance.
The exact phrasing and scoring of behavioral anchors for each criterion with examples are presented in Tables \ref{tab:rubric_feedback_p1} and \ref{tab:rubric_feedback_p2}

The predictive validity of these dimensions is underscored by their associations with observed trainee responses. As shown in Figure~\ref{fig:quality_associations}, Urgent feedback was strongly linked to immediate behavioral adjustments, whereas Reflective prompts were more likely to elicit verbal acknowledgment rather than action. These divergent associations suggest that different feedback types may selectively activate cognitive or behavioral processing in trainees. Some findings were counterintuitive: notably, higher levels of Encouragement were associated with a lower likelihood of both \revT{behavioral adjustment} and verbal acknowledgment \rev{from the trainee}. This likely reflects the role of encouragement as positive reinforcement—often used to affirm correct performance—thus requiring neither additional adjustment nor a verbal response. Similarly, \emph{Clarity} was associated with increased verbal acknowledgment but showed no significant link to \revT{behavioral adjustment}. This pattern may be due to the orthogonality between clarity and actionability: a statement can be easy to understand without necessarily implying that action is needed. Clear feedback may also lower the threshold for verbal response, making it easier for trainees to affirm receipt. Finally, Figure~\ref{fig:quality_wordclouds} highlights representative linguistic patterns across criteria, further supporting their face validity and offering actionable insights for feedback design. The emergence of these quality dimensions demonstrates the potential of AI-derived labels to structure, assess, and ultimately improve surgical feedback practices in real-world settings.

Our framework is the first to combine LLM-driven discovery of feedback quality dimensions with scalable rating via a behaviorally grounded rubric. Unlike prior methods, we do not rely on predefined rubrics or rigid taxonomies \citep{zheng2023judging}. Instead, we task the LLM with discovering evaluative criteria from real surgical interactions and clinically grounded knowledge, enabling the model to surface delivery qualities relevant to actual training dynamics (i.e., what makes a good feedback). These criteria are defined in plain, human interpretable language, and scored on a Behaviorally Anchored Rating Scale (BARS), a well-established tool in psychometrics that enhances human interpretability and support Human-AI alignment through representative behavioral anchors \citep{holland2022reliability}. This allows both humans and AI systems to consistently score feedback based on the same standards. \rev{Explicit anchoring also constrains the influence of a model's internalized priors. When a rating construct is left underspecified, an LLM annotator falls back on its own internalized notion of the concept; additional prompt detail corrects only a fraction of the resulting errors, and misaligned definitions shift ratings without any corresponding drop in the model's reported confidence~\citep{casanova2026limits}. Behavioral anchors narrow that interpretive room, and the human calibration study provides a direct check that the narrowing worked.}

Technically, our approach diverges from traditional unsupervised clustering \citep{liu2022stability}, topic modeling \citep{grootendorst2022bertopic}, or keyword-based methods \citep{tausczik2010psychological} by focusing on delivery quality rather than content similarity.

Our method advances prior \emph{``LLM‑as‑a‑judge''} work \citep{gu2024survey} in two fundamental ways.  
First, we shift the role of the LLM from \emph{applying} a fixed rubric to \emph{discovering} that rubric de novo. We encourage completeness of the rubric by \rev{drawing five independent candidate rubrics} at a high sampling temperature (encouraging generation diversity \citep{patel2024exploring}), each exposed to a different subsample of real feedback\rev{, and then aggregating them. The consolidated rubric exceeds the mean of the individual rubrics on all four outcomes and the best individual rubric on three of the four (Table~\ref{tab:multiagent_auc}). This procedure surfaces behaviorally salient qualities that a single, low-temperature draw fails to surface, such as} \emph{Timeliness} \rev{and} \emph{Urgency}\rev{.} Second, we solve the discovery stability problem by passing the diverse candidate set through a deterministic consolidation stage. Hierarchical clustering on scoring correlations identifies semantically overlapping criteria, and the LLM then produces a finalized phrasing and BARS anchors for each cluster. Across five seeds, the final six‑dimension rubric exhibits negligible lexical drift while retaining the conceptual breadth unlocked during the brainstorming step.

Expressing every criterion in plain language and anchoring it with illustrative examples enables dual use: the same rubric can be parsed reproducibly by an LLM at scale and interpreted reliably by human raters with minimal calibration.  This unification of discovery, formalization and scoring distinguishes our pipeline from unsupervised topic models—whose latent factors are not scorable \citep{chang2009reading}—and from black‑box classifiers, whose decision rationales remain opaque \citep{lipton2018mythos, ribeiro2016should}. By addressing \emph{completeness}, \emph{stability} and \emph{interpretability} in a single workflow, we provide a principled path for deploying LLMs in safety‑critical, clinician‑facing settings. \rev{Anchoring each criterion to a specific, observable behavior rather than to a broad construct is also what makes the rubric measurable: in LLM psychometrics, instruments tied to a particular behavior track model behavior far more closely than coarse dispositional scales do, and that correspondence degrades precisely where context dominates the judgment~\citep{kocielnik2026psychometric}.%
% Optional companion citation -- delete the next \citep and the preceding
% comment if a single reference reads better here:
{} The same divergence can be traced to distinct internal representations for what a model reports and what it does~\citep{kocielnik2026representational}.}

We rigorously evaluated the utility of our discovered quality criteria across multiple dimensions. First, the criteria independently predicted trainee behavioral and verbal outcomes as well as follow-up trainer reactions better than prior topic-based and human-annotated baselines, showing 3–12\% performance improvements. Second, we confirmed the criteria's interpretability in a human‑rating study: human raters with clinical knowledge, applied the rubric to real feedback examples and achieved substantial agreement—both with each other and with an LLM on five of the six dimensions (quadratic $\kappa$ = 0.60–0.70); the remaining dimension showed moderate agreement. Third, our analysis uncovered meaningful associations between specific feedback qualities and subsequent trainee behavior. For instance, timely and actionable feedback strongly predicted behavior change, while reflective and clear feedback was more likely to prompt verbal acknowledgment. These findings confirm the practical validity of our dimensions and underscore their clinical relevance. \rev{The associations also support the rubric's relevance to immediate feedback uptake; whether rubric-guided feedback improves longer-term learning, surgical skill, or patient outcomes requires prospective evaluation.}
Our framework delivers a scalable, transparent way to quantify intra‑operative feedback quality—communication that shapes surgical trainee learning curves \citep{bonrath2015comprehensive, ma2022tailored} and ultimately impacts patient outcomes and safety \citep{d2020evaluating, greenberg2021association}. In practice, the rubric can underpin point‑of‑care dashboards that highlight especially actionable or unclear coaching, longitudinal curriculum analytics that flag faculty‑wide gaps, and automated quality‑assurance audits that monitor for safety‑critical patterns such as high urgency coupled with low clarity. \rev{The same scores can serve a generative role: systems that draft surgical feedback automatically are currently evaluated mainly on whether the content is correct~\citep{nasriddinov2026generating}, and a delivery-quality rubric supplies the complementary signal needed to choose among candidate phrasings of the same instruction.} Because each criterion is expressed in plain language yet is automatically scorable by a large language model, the tool supports human‑in‑the‑loop deployment: clinicians can inspect the rubric, contest low scores, and review exemplar anchors, ensuring both accountability and trust. \rev{The Mistral and Qwen evaluations show that the finalized rubric can be applied beyond GPT-4o with comparable downstream predictive performance, although the models agreed with GPT-4o only moderately on individual scores. Their openly available weights provide a route to institutional self-hosting of scoring, offering greater control over data handling and model versioning. Agreement was also not uniform across criteria: it was substantial for \emph{Actionability} and \emph{Urgency} but considerably lower for \emph{Encouragement} and \emph{Timeliness} (Table~\ref{tab:cross_model}B). This pattern is consistent with models differing in their internalized definition of a construct rather than in their ability to follow a rubric~\citep{casanova2026limits}, and argues for re-verifying rubric application against human raters whenever the scoring model is changed, rather than assuming transfer.}

% Limitaitons
Yet, several limitations warrant mention. First, our analysis relied solely on text; prosodic cues in audio, instrument motion \rev{from} video, and contextual OR events were not modeled and could modulate how feedback is interpreted by trainees. Second, our evaluation was retrospective \rev{and drew on a single intraoperative dataset, although that dataset spans several urologic procedure types across different organs and multiple trainers and trainees~\cite{wong2023development}. Because consolidation used scoring patterns computed across that corpus, utterances later held out in cross-validation contributed to defining the rubric, even though development drew only on the feedback text while evaluation drew on separately annotated trainee and trainer behavior observed outside it. Preprocessing that uses no outcome labels can nonetheless bias cross-validation estimates in either direction~\citep{moscovich2022crossvalidation}, and the reported values should therefore be read as conditional on a corpus-derived rubric. Independent external validation across institutions and clinical settings is needed to establish transferability.} A prospective study that provides real‑time rubric scores to trainers—and measures downstream \revT{behavioral adjustment}—will be an important next step that our lightweight, text‑only pipeline readily supports. \rev{Third, we did not independently assess trainees' perspectives; observable responses may reflect compliance rather than agreement or learning, particularly within hierarchical training environments.} Finally, \revT{rubric‑discovery} relied on GPT‑4o, a proprietary model that may evolve over time. \rev{We demonstrated application of the finalized rubric by human clinical raters and two open-weight models, Mistral Small 3.2 and Qwen3-Next 80B-A3B, and publish the rubric, BARS anchors, and scoring prompt to support reproducible implementation and reduce dependence on proprietary models for scoring.} \revDel{To reduce vendor lock‑in we evaluated the application of the discovered rubric with human raters with clinical knowledge and we publish the final six‑dimension rubric, BARS anchors and scoring prompt, enabling re‑implementation with open‑source LLMs or future clinical language models.}

% Broader impact
Although developed for the operating room, our discover‑and‑score pipeline can generalize to any clinical setting that relies on concise spoken guidance—such as ICU rounds, nursing shift reports, or telemedicine consultations. Because the model operates on plain text, a small speech‑to‑text transcript is all that must be transmitted; bulky audio or video files are unnecessary. This lightweight footprint enables remote mentorship and feedback‑quality monitoring in bandwidth‑constrained settings, with the potential to narrow global disparities in surgical training and, ultimately, patient outcomes. Furthermore applying our discover-and-score pipeline to other clinical domains—multidisciplinary tumor boards, emergency‑department hand‑offs, or virtual rehabilitation sessions—could yield domain‑specific communication metrics that remain interpretable to frontline staff. Linking feedback‑quality scores directly to downstream training or quality‑of‑care indicators would close the loop between communication analysis and measurable performance improvement, advancing the integration of transparent AI assistants into everyday digital‑health workflows.

\section{Methods}
\label{sec:methods}

\subsection{Ethics Approval}
This study utilized datasets collected in accordance with strict ethical guidelines and approved by the Institutional Review Board (IRB) at the University of Southern California (HS-17-00113). All participants provided written informed consent prior to data collection. To ensure participant privacy and confidentiality, all datasets were de-identified before any model development or analysis was conducted.

\subsection{Surgical Feedback Dataset}
%% Feedback-Frequency Table (compact width) %%
\begin{table}[t]
\centering
\small
\begin{tabular}{p{1.8cm}lrrrr}
\toprule
\textbf{Category} & \textbf{Dimension} & \textbf{Count} & \textbf{Freq} & \textbf{Count/Case} & \textbf{Words/Line} \\
\midrule
Feedback & Instances & 4210 & 100.0\% & $131.6\pm77.6$ & $8.3\pm6.9$ \\[2pt]
\midrule
\multirow{2}{1.8cm}{Trainee Behavior}
  & Verbal Ack. & 1944 & 46.2\% & $60.8\pm31.7$ & $10.2\pm7.5$ \\[2pt]
  & Behavioral Adj. & 1866 & 44.3\% & $58.3\pm41.1$ & $8.4\pm6.5$  \\[2pt]
\midrule
\multirow{2}{1.8cm}{Trainer Reaction}
  & Approval    & 619 & 14.7\% & $19.3\pm18.4$ & $9.1\pm6.9$ \\[2pt]
  & Disapproval &  85 &  2.0\% &  $2.7\pm2.5$  & $9.4\pm6.5$ \\
\bottomrule
\end{tabular}
\caption{Statistics of behavior categories in our dataset. We report absolute counts (\textit{Count}), relative frequency across all feedback instances (\textit{Freq}), prevalence per surgical case (\textit{Count/Case}), and mean words per utterance with standard deviation (\textit{Words/Line}).}
\label{tab:dataset-stats}
\end{table}

We used a dataset of real-world intraoperative feedback collected during robot-assisted surgeries, as introduced by Wong et al. \cite{wong2023development}. Audio was recorded via wireless microphones worn by the surgical team, and synchronized endoscopic video was captured from the da Vinci Xi surgical system \cite{freschi2013technical}, providing a first-person surgical view. Using an external recording setup, audio and video streams were aligned and stored.

Utterances constituting surgical feedback—defined as trainer statements intended to modify trainee thinking or behavior—were manually identified and transcribed by surgical residents. Only utterances meeting this definition were included; other conversational content was excluded. The resulting dataset includes 4,210 feedback instances (Table \ref{tab:dataset-stats}).

Each instance was further annotated for two categories of behavioral outcomes. \emph{Trainee Behavior} annotations captured whether the trainee responded to the feedback, including: (i) \emph{Verbal Acknowledgment}, defined as a verbal or audible reaction confirming that the feedback was heard, and (ii) \emph{Behavioral Adjustment}, defined as a behavioral change directly corresponding to the preceding feedback. \emph{Trainer Reaction} annotations captured the trainer’s response to the observed trainee behavior: (i) \emph{Approval}, where the trainer verbally indicated satisfaction with the trainee's response, and (ii) \emph{Disapproval}, where the trainer verbally demonstrated that they were not yet satisfied with the observed trainee behavioral change. The frequency, per-case prevalence, and average feedback length for each dimension are summarized in Table \ref{tab:dataset-stats}. All annotation procedures followed standardized guidelines, and further details are available in the original dataset publication \cite{wong2023development}.

\subsection{Automated Discovery of Feedback Quality Criteria}
\label{sec:ai_discovery}

Our AI-based framework extracts interpretable feedback quality dimensions from surgical training data, designed to be scorable on a Behaviorally Anchored Rating Scale (BARS)~\citep{schwab1975behaviorally, jacobs1980expectations}. Inspired by prior tools for surgical skill assessment, such as OSATS~\citep{van2010objective}, EASE~\citep{haque2022assessment}, and DART~\citep{vanstrum2021development}, these criteria facilitate both human interpretation and automated scoring using LLMs. An overview is shown in Figure~\ref{fig:task_overview}.

\paragraph{Domain-Guided Initiation}

We initiate criteria discovery by injecting domain-specific knowledge into the prompt. Following in-context learning~\citep{kojima2022large}, GPT-4o is provided with formal definitions of key surgical feedback outcomes~\citep{wong2023development}—\textit{Behavioral Adjustment}, \textit{Verbal Acknowledgment}, and \textit{Trainer Approval}—as well as a definition of feedback: \textit{``Dialogue intended to modify trainee thinking or behavior.''} These definitions are accompanied by 50 randomly selected unlabeled feedback examples ($<$1\% of the dataset), enabling the LLM to infer relevant evaluative dimensions through analogical reasoning~\citep{ozturkler2022thinksum, wang2022rationale}. Although LLMs are pre-trained with general language capabilities \citep{jiang2023large, wei2021finetuned}, recent work indicates that the domain-specific information provided during prompting can help LLMs perform better by allowing them to reason appropriately within the context of the task \citep{maharjan2024openmedlm}. Domain-specific information is especially important in specialized domains such as clinical natural language processing \citep{sivarajkumar2024empirical}.

\paragraph{\revT{Multi-Instance} Criteria Generation}
\label{sec:criteria_discovery}

\rev{We used five independent GPT-4o sessions with the same discovery instructions, each receiving a different randomly sampled subset of 50 unlabeled feedback instances. These sessions used the same pretrained model without separate training or direct communication between sessions. Their candidate criteria were subsequently combined through the scoring-based clustering and wording consolidation procedure described below.} \revDel{We implemented a multi-agent setup where five GPT-4o instances independently propose candidate criteria. Each agent received a distinct random subset of 50 unlabeled feedback samples and was prompted to identify generalizable, abstract quality dimensions predictive of the defined outcomes.} To encourage creative diversity, \revT{instances} were configured with a high temperature ($T=1.0$), which promotes variability in outputs---a key benefit in exploratory tasks like criteria discovery. Prior work has shown that higher temperatures enhance idea generation and reduce output homogenization, albeit at the cost of reduced determinism~\citep{windisch2024impact, anderson2024homogenization}. Here, we prioritized discovering novel feedback qualities over reproducibility.

\rev{Multi-instance prompting here denotes repeated independent sampling. The five sessions are stateless and do not communicate, so each is an independent draw from the same pretrained model under identical instructions, varying only in its 50-example subsample and the sampling temperature. Diversity across draws therefore arises from sampling, and the consolidation step described below aggregates the draws into a single rubric.}

Critically, the prompt specified that each proposed dimension should: (1) be \emph{definable in abstract terms}, meaning it must describe a generalizable quality applicable across feedback lines (rather than context-specific actions or examples), and (2) be feasibly scorable on a 5-point BARS scale based solely on a single transcribed feedback line, without access to surrounding dialogue or video context. Each \revT{instance} was instructed to include behavioral anchors for three key levels: scores of 1, 3, and 5. These anchors served as representative examples for raters to interpret the quality being described. Each \revT{instance} was asked to format its output in a structured tabular layout, listing the dimension name, its definition, and descriptions or examples of what constitutes a score of 1, 3, and 5. This structured prompting strategy ensured consistency across generations and interpretability of the resulting criteria. The prompt wording is provided in \rev{Appendix~\ref{apx:llm_prompts}} as \textit{``Prompt Template for Quality-Criteria Discovery''}.

\paragraph{Criteria Consolidation and Definition Phrasing Finalization}

We applied each \revT{instance's} criteria to the full feedback dataset and computed a Spearman correlation matrix across all discovered dimensions. Hierarchical clustering (single linkage, Euclidean distance) revealed convergence patterns across \revT{instances}. \rev{The resulting correlation matrix and dendrogram are shown in Appendix~\ref{apx:consolidation_clustering}, Figure~\ref{fig:criteria_clustermap}.} For cluster‐number selection, we cut the dendrogram at successive values of $k$ and computed the mean silhouette coefficient for each cut; the peak silhouette value occurred at $k=6$, indicating six well-separated, internally cohesive clusters of semantically similar criteria \citep{rousseeuw1987silhouettes}. For each cluster, GPT-4o (in deterministic mode, $T = 0.0$) synthesized a unified dimension definition to ensure reproducibility following \cite{windisch2024impact, patel2024exploring}. Prompts emphasized non-overlap, clarity, and applicability to isolated transcribed feedback, producing a refined set of six interpretable and domain-relevant feedback quality dimensions \citep{patel2024exploring}. The prompt used is provided in \rev{Appendix~\ref{apx:llm_prompts}} as \textit{``Prompt Template for Criteria Consolidation Phrasing per Cluster''}. \rev{Rubric discovery and consolidation were performed once, before the predictive splits, using unlabeled feedback text and no behavioral outcome labels. Candidate criteria were scored across the full corpus and clustered by the correlation structure of those scores, with the number of clusters selected by mean silhouette coefficient, an internal clustering criterion~\citep{rousseeuw1987silhouettes}. The finalized rubric is therefore derived from the corpus as a whole, and was then held fixed across all predictive splits and behavioral outcomes.}

Wording stability after five repetitions of the consolidation step was quantified with a cross‑seed cosine‑distance metric: each rubric definition was embedded using the \texttt{all‑MiniLM‑L6‑v2} sentence‑transformer \citep{sentence28:online}, pair‑wise cosine distances ($1-\cos \theta$) were computed between definitions that shared the same cluster index but originated from different seeds, and the resulting values were averaged.  Distances $\le 0.05$ are widely used as the near‑duplicate threshold in large‑scale text‑deduplication pipelines \citep{mishra2020similarity,rodier2020online}; our mean distance of 0.02 therefore indicates negligible lexical drift.  To evaluate whether conceptual breadth was retained, we tokenized each definition into uni‑, bi‑ and tri‑grams, embedded every term, and deemed a brainstorming term ``covered'' if its embedding showed cosine similarity $\ge 0.80$ with any term in the consolidated rubric. This 0.80 cut‑off is consistent with thresholds employed for near‑duplicate detection and semantic‑match evaluation in recent clinical‑NLP studies \citep{tumre2025improved,zhao2024x}.  Under this criterion, the final six‑dimension rubric covered 61.3\,\% of the vocabulary introduced during brainstorming, demonstrating that consolidation preserved the majority of the original conceptual space while standardizing phrasing.

To verify that the unified six-dimension rubric retained (or improved upon) the predictive signal discovered by individual \revT{instances}, we first applied each set of criteria to every feedback instance using LLM-as-a-judge approach detailed in \Sref{sec:llm-as-a-judge}, yielding a vector of 5-point ordinal ratings—one score per dimension. These ratings were then used as predictors in a logistic-regression model for each behavioral outcome. \rev{For each behavioral outcome, we fitted a separate logistic-regression model for each instance's criteria scores and an additional model using the consolidated six-dimension rubric scores (Consolidated Criteria). GPT-4o was used exclusively for prompt-based inference; no LLM fine-tuning or parameter updates were performed.} Model performance was evaluated over three independent, stratified 80/20 train–test splits generated with three fixed random seeds. As summarized in Table \ref{tab:multiagent_auc}, the consolidated rubric achieved the highest mean AUC across three of the four outcomes, with an above average AUC for the remaining outcome. \rev{Because behavioral outcome labels are unavailable during discovery, the strongest individual rubric cannot be identified in advance, so aggregation across draws rather than selection among them determines the rubric carried forward.} These findings empirically support the hierarchical clustering and synthesis step, demonstrating that consolidation not only harmonizes terminology but also concentrates predictive signal for downstream modelling.

\begin{table}[!ht]
\centering
\caption{Predictive power of quality–score criteria produced by each GPT-4o individual \revT{instance} versus the final consolidated score (``Consolidated Criteria''). Values are mean AUC (± SD) across the three random-seed folds. Highest AUC for each outcome is bold-faced.}
\label{tab:multiagent_auc}
\begin{tabular}{lcccc}
\toprule
 & \multicolumn{2}{c}{\shortstack{\textbf{Trainee Reaction} \\ \textbf{to Feedback}}} & \multicolumn{2}{c}{\shortstack{\textbf{Trainer Reaction} \\ \textbf{to Trainee Reaction}}}\\
\cmidrule(lr){2-3}\cmidrule(lr){4-5}
\textbf{Quality Rubric source} &
\textbf{\revT{Behavioral Adj.}} &
\textbf{Verbal Ack.} &
\textbf{Approval} &
\textbf{Disapproval}\\
\midrule
GPT-4o \revT{Instance} \#1 & $0.66\pm0.01$ & $0.62\pm0.02$ & $0.63\pm0.01$ & $0.58\pm0.06$\\
GPT-4o \revT{Instance} \#2 & $0.67\pm0.01$ & $0.63\pm0.02$ & $0.65\pm0.01$ & $\mathbf{0.64}\pm0.02$\\
GPT-4o \revT{Instance} \#3 & $0.73\pm0.02$ & $0.68\pm0.01$ & $0.63\pm0.01$ & $0.62\pm0.01$\\
GPT-4o \revT{Instance} \#4 & $0.71\pm0.01$ & $0.67\pm0.02$ & $0.63\pm0.01$ & $0.63\pm0.04$\\
GPT-4o \revT{Instance} \#5 & $0.73\pm0.03$ & $0.64\pm0.01$ & $0.63\pm0.01$ & $0.62\pm0.06$\\
\midrule
\rev{Mean of Instances \#1--5} & \rev{$0.70$} & \rev{$0.65$} & \rev{$0.63$} & \rev{$0.62$}\\
\midrule
\textbf{Consolidated Criteria} & $\mathbf{0.74}\pm0.01$ & $\mathbf{0.71}\pm0.02$ & $\mathbf{0.66}\pm0.02$ & $0.63\pm0.01$\\
\bottomrule
\end{tabular}
\end{table}

\subsection{Automated Feedback Scoring Based on Discovered Criteria}
\label{sec:ai_scoring}

To systematically assess surgical feedback quality at scale, we employed a large language model (GPT-4o) to score real-world transcribed feedback instances using the rubric developed through our discovery process (Tables~\ref{tab:rubric_feedback_p1} and~\ref{tab:rubric_feedback_p2}). This process follows the \emph{LLM-as-a-judge} paradigm~\citep{gu2024survey, li2024generation}, where the language model acts as a consistent evaluator applying structured criteria.

\paragraph{Input for Scoring}

Each feedback instance was independently annotated by GPT-4o, which was prompted with (a) the full set of six scoring criteria along with their definitions and representative examples (Tables \ref{tab:rubric_feedback_p1}, \ref{tab:rubric_feedback_p2}), and (b) the specific feedback text to be rated. This structured input format enabled the model to interpret and apply the rubric definitions grounded in a behaviorally anchored rating scale (BARS). The prompt structure used for this task is detailed in \rev{Appendix~\ref{apx:llm_prompts}} as \textit{``Prompt Template for Multi-Criteria Feedback Scoring''}.
%~\ref{tab:surgicalFeedbackScoringPrompt}.

\paragraph{LLM-as-a-judge Scoring}
\label{sec:llm-as-a-judge}

The LLM assigned a score from 1 to 5 for each of the six criteria, returning a structured list of numerical ratings per feedback instance. All instances were scored individually in separate API calls, without batching, to minimize potential cross-instance contamination and data leakage~\citep{schroeder2024can}. This approach is consistent with recent methodological best practices in judgment elicitation with LLMs~\citep{zheng2023judging}.

To ensure reproducibility and reduce variance in scoring, we set the model’s temperature to 0.0, encouraging deterministic outputs. To evaluate scoring consistency, we conducted a repeated annotation of each feedback instance using the same model and prompt configuration, enabling calculation of inter-rater agreement for each quality dimension.

\paragraph{Human-AI Alignment Calibration}
\label{sec:method_scoring_hum_ai_align}

To assess the alignment between LLM-based and human judgment, we conducted a calibration study on a stratified sample of 30 LLM-scored feedback instances. For each of the six discovered feedback quality dimensions, we selected five examples spanning the full BARS scoring spectrum: two instances with high scores (4 or 5), one with a mid-range score (3), and two with low scores (1 or 2). This stratification ensured representative coverage across the rating scale for every dimension.

Two human raters with surgical domain knowledge independently evaluated these instances, using the original LLM-discovered definitions and applying the same 5-point BARS scoring system. Raters were blinded to the LLM-generated scores to prevent bias. Following the initial rating phase, any disagreements of two or more points on the BARS scale were discussed collaboratively, allowing the raters to reconcile interpretations and establish a consensus score for each instance.

Importantly, in cases where both human raters independently agreed on a score that differed from the LLM's original rating, these instances were incorporated as new illustrative examples into the scoring rubric (up to two new examples per anchor). This iterative grounding process reinforced the behavioral anchoring of each quality dimension and aligned with rubric refinement practices recommended by prior work on human-AI collaboration in alignment tasks~\citep{pan2024human}.

\subsection{Comparison to Existing Automated Criteria Extraction}
\label{sec:prior_comparison}

Most prior approaches to discovering evaluative dimensions from text rely on unsupervised topic modeling (e.g., LDA, BERTopic) or black-box LLM scoring frameworks like LLM-as-a-judge. While topic models can surface latent themes, they do not yield actionable, interpretable evaluation criteria aligned with domain-specific outcomes such as behavioral adjustment or trainer approval. Similarly, LLM-based scorers often replicate pre-defined preferences or rubrics without uncovering novel dimensions. Moreover, these systems typically do not provide explanations for their decisions unless paired with post hoc interpretability methods that generate human-understandable rationales or explanations \citep{mosca2022shap}. These methods fall short in safety-critical domains like surgical education, where human-verifiable, domain-grounded criteria are essential for both evaluation and training.

Our method addresses this gap through a two-stage, LLM-assisted pipeline. In the first stage, we initiate domain-grounded criteria discovery using a \revT{multi-instance} prompting strategy, where separate LLM \revT{instances} generate candidate evaluation criteria informed by clinical definitions and feedback examples. These outputs are then consolidated into a set of interpretable, BARS-compatible rating dimensions. In the second stage, we leverage these discovered criteria to score new feedback instances using a structured LLM-as-a-judge approach. This enables transparent, repeatable evaluation aligned with domain outcomes, bridging the strengths of human-centered scale development and LLM-based automation. A comparison of the properties of our method against prior approaches is presented in Table~\ref{tab:comparison_methods}, illustrating its unique ability to support interpretable discovery, domain-grounded scoring, and dual usability by both humans and LLMs.

\begin{table}[h!]
\centering
\caption{\textbf{Comparison of existing methods for automated evaluation dimension discovery and scoring.} Our method uniquely combines interpretable criteria discovery with scoreability and dual usability by humans and LLMs. Unlike topic modeling or direct LLM scoring approaches, it enables domain-grounded, BARS-compatible dimension induction and structured evaluation.}
\begin{tabular}{lccccc}
\hline
\textbf{Method} & \textbf{Discovers} & \textbf{Scoreable} & \textbf{Domain} & \textbf{Inter-} & \textbf{Human +} \\
                & \textbf{Dimensions} & \textbf{Criteria} & \textbf{Grounded} & \textbf{pretable} & \textbf{LLM Scoring} \\
\hline
LDA / BERTopic         & \checkmark & \xmark & \xmark & \textcolor{orange}{Ambiguous} & \xmark \\
LLM-as-a-Judge         & \xmark     & \checkmark  & \textcolor{orange}{Tuned} & \xmark & \textcolor{orange}{LLM-only} \\
\textbf{Our Method}    & \checkmark & \checkmark & \checkmark & \checkmark & \checkmark \\
\hline
\end{tabular}
\label{tab:comparison_methods}
\end{table}

\subsection{Predictive Modeling of Behavioral Outcomes}
\label{sec:evaluation_setup}

We evaluated the predictive utility of LLM-discovered feedback quality criteria by modeling four behavioral outcomes annotated by human experts in \cite{wong2023development}: \emph{Trainee \revT{Behavioral Adjustment}}, \emph{\revT{Trainee Verbal Acknowledgment}}, \emph{Trainer Approval}, and \emph{Trainer Disapproval}. \rev{We selected Random Forest classifiers to accommodate potentially nonlinear associations and interactions among feedback quality scores and previously published feedback categories without prespecifying their functional forms~\citep{breiman2001random}. Using the same classifier family and evaluation protocol across feature configurations supported a consistent comparison of their predictive utility.} We implemented a fivefold \textit{stratified} cross-validation procedure to ensure that each fold preserved the original class distribution for the respective outcome. Within each training fold, we performed nested hyperparameter tuning using an inner fivefold stratified cross-validation loop. Random Forest classifiers were tuned via grid search over the following parameter ranges: number of estimators \texttt{[200, 300, 400, 500, 1000]}, maximum number of features \texttt{[10, 25, 50]}, maximum tree depth \texttt{[20, 50]}, and minimum samples per leaf \texttt{[5, 20]}. The Gini impurity index was used as the splitting criterion, and the best hyperparameters were selected based on AUROC performance on the validation folds. To further address the effects of class imbalance during model training—particularly for less frequent outcomes such as \emph{Trainer Disapproval}—we applied class weighting using King’s method~\cite{king2001logistic}, which adjusts model estimation to reduce bias in rare event prediction.

We compared models using three types of features: (1) our AI-derived quality scores, (2) previously published topic modeling features~\cite{kocielnik2024human}, and (3) previously published human-defined feedback categories~\cite{wong2023development}, both individually and in combination. Feature matrices were constructed accordingly, and the same cross-validation protocol was applied across all feature configurations.

Each fold’s predictions were evaluated using standard classification metrics, including Area Under the Receiver Operating Characteristic Curve (AUROC), accuracy, precision, and recall. We report AUROC with 95\% confidence intervals calculated using the DeLong method \citep{sun2014fast} over the pooled held-out predictions. Model performance across feature configurations and behavioral outcomes is presented in Table~\ref{tab:predicting_outcomes}.

% NEW (R2): multicollinearity check + open-weight model replication
\rev{We assessed relationships among the six finalized rubric scores using Spearman correlations and variance inflation factors (VIFs). To evaluate application beyond GPT-4o, Mistral Small 3.2 (24B) and Qwen3-Next 80B-A3B scored the same corpus using the finalized rubric and scoring prompt at temperature 0. Agreement with GPT-4o was calculated on matched instances using quadratic-weighted $\kappa$ for each criterion, their arithmetic mean, and the proportion of scores within one scale point. Agreement confidence intervals used 2,000 percentile-bootstrap resamples of surgical cases, retaining all utterances and six scores together and recalculating mean $\kappa$ in each resample.}

% NEW (R2): case-grouped vs. outcome-stratified CV sensitivity analysis
\rev{For all three models, an additional sensitivity analysis compared fivefold outcome-stratified and case-grouped cross-validation for predicting \emph{Behavioral Adjustment} and \emph{Verbal Acknowledgment}, using a fixed 500-tree Random Forest with the six scores as predictors. Grouped folds kept all utterances from each surgical case together. AUROC confidence intervals used 1,000 percentile-bootstrap resamples of individual instances from pooled out-of-fold predictions.}

% Table DeLong's Z-test + 95% CI for dull and reduced models
\begin{table*}[t!]
\centering
\footnotesize{
% NOTE: this was a nicematrix NiceTabular with the [colortbl-like] option.
% That key was removed in recent nicematrix releases (it is now the default
% behaviour), which made the document fail to compile on up-to-date TeX Live.
% Nothing nicematrix-specific was actually used here -- the \CodeBefore block
% was empty and all shading is plain colortbl -- so this is now a plain
% tabular, which is immune to that package's version churn.
% Column widths: the shaded Full Model columns are given explicit widths so the
% bold "AUC" header cannot overflow and be sliced by the next column's
% \columncolor box. The dAUC columns are right-aligned and the CI columns
% left-aligned so that bold (significant) and regular (non-significant) values
% still line up with each other -- centring them made the unbolded rows drift.
\begin{tabular}{@{} p{1.14in} % Outcome Name
>{\columncolor{Gray}\centering\arraybackslash}p{0.34in} @{\hskip 0.03in} % Full AUC
>{\columncolor{Gray}\centering\arraybackslash}p{0.58in} @{\hskip 0.06in} % Full CI
r @{\hskip 0.04in} % ΔAUC No AI
l @{\hskip 0.06in} % CI No AI
r @{\hskip 0.04in} % ΔAUC No Human
l @{} % CI No Human
}
\toprule
\textbf{Behavior Outcome} & \multicolumn{2}{c}{\textbf{Full Model}} & \multicolumn{2}{c}{\textbf{Without AI Quality}} & \multicolumn{2}{c}{\textbf{Without Prior}} \\
& \textbf{AUC} & \textbf{95\% CI} & \multicolumn{1}{c}{\textbf{$\Delta$AUC}} & \multicolumn{1}{c}{\textbf{95\% CI}} & \multicolumn{1}{c}{\textbf{$\Delta$AUC}} & \multicolumn{1}{c}{\textbf{95\% CI}} \\
\midrule

\rowcolor{Gray}\multicolumn{7}{l}{\textbf{Compared to Topic Modeling ~\cite{kocielnik2024human}}} \\
\midrule
Beh. Adjustment & 0.77 & (0.76, 0.79) & $\mathbf{-0.080}^{*}$ & $(-0.093, -0.066)$ & $\mathbf{-0.021}^{*}$ & $(-0.029, -0.013)$ \\

\revT{Verb. Acknowledgment} & 0.73 & (0.71, 0.74) & $\mathbf{-0.061}^{*}$ & $(-0.074, -0.048)$ & $\mathbf{-0.016}^{*}$ & $(-0.025, -0.006)$ \\

Trainer Approval & 0.69 & (0.67, 0.71) & $\mathbf{-0.017}^{\dagger}$ & $(-0.035, -0.000)$ & $\mathbf{-0.026}^{*}$ & $(-0.040, -0.011)$ \\

Trainer Disapproval & 0.63 & (0.58, 0.69) & $\mathbf{-0.065}^{*}$ & $(-0.127, -0.004)$ & $-0.031\phantom{^{*}}$ & $(-0.086, 0.023)$ \\

\midrule
\rowcolor{Gray}\multicolumn{7}{l}{\textbf{Compared to Human Categories ~\cite{wong2023development}}} \\
\midrule
Beh. Adjustment & 0.78 & (0.77, 0.80) & $\mathbf{-0.082}^{*}$ & $(-0.096, -0.067)$ & $\mathbf{-0.032}^{*}$ & $(-0.039, -0.024)$ \\

\revT{Verb. Acknowledgment} & 0.74 & (0.73, 0.76) & $\mathbf{-0.064}^{*}$ & $(-0.078, -0.051)$ & $\mathbf{-0.032}^{*}$ & $(-0.043, -0.022)$ \\

Trainer Approval & 0.69 & (0.67, 0.71) & $\mathbf{-0.058}^{*}$ & $(-0.079, -0.036)$ & $\mathbf{-0.026}^{*}$ & $(-0.041, -0.011)$ \\

Trainer Disapproval & 0.62 & (0.56, 0.68) & $-0.026\phantom{^{*}}$ & $(-0.085, 0.033)$ & $-0.017\phantom{^{*}}$ & $(-0.065, 0.032)$ \\

\bottomrule
\end{tabular}
}
\vspace{3pt}
\caption{AUROC values for models using both \textbf{AI Quality Ratings} and \textbf{Prior Work Categories}, and the respective drops in AUROC when either feature set is removed. Rows are grouped by comparison baseline. \rev{A drop is marked $^{*}$ when the entire 95\% CI for $\Delta$AUC lies below 0, that is, when its upper bound is strictly less than 0. $\dagger$ marks a marginal drop whose upper bound lies below 0 by less than 0.001. Both $\Delta$AUC columns and their intervals are reported to three decimals because several bounds lie close to 0 and two-decimal rounding would not distinguish a bound of $-0.004$ from one of 0; absolute AUC values are reported to two decimals.}}
\label{tab:delong_comparative_analysis}
\vspace{-0.5em}
\end{table*}

\subsection{Model Comparison and Statistical Significance Testing}
\label{sec:delong_comparison}

To assess the independent predictive contributions of AI-derived quality scores and previously published feedback annotations, we conducted a comparative analysis of model performance using DeLong’s test for correlated receiver operating characteristic (ROC) curves. Specifically, we built three models for each behavioral outcome: (1) a \emph{full model} incorporating both AI-derived quality scores and prior annotation features (topic modeling~\cite{kocielnik2024human} and human-defined categories~\cite{wong2023development}), (2) a \emph{reduced model without AI quality scores}, and (3) a \emph{reduced model without prior annotation features}.

All models were trained using Random‑Forest classifiers with class‑weight balancing to account for outcome imbalance. Hyper‑parameters were tuned by nested cross‑validation: a fivefold stratified outer loop estimated generalization performance, while a threefold stratified inner loop executed a grid search over the number of trees \texttt{[200, 300, 400, 500, 1000]}, the number of features considered at each split \texttt{[10, 25, 50]}, the maximum tree depth \texttt{[20, 50]}, and the minimum samples per leaf \texttt{[5, 20]}. The hyper‑parameter set that maximized AUROC within the inner loop was refit on the full outer‑training fold and then evaluated on its corresponding held‑out outer‑test fold. AUROC scores were computed on these outer‑test sets, and DeLong’s method provided 95\,\% confidence intervals (CIs) both for each individual AUROC and for the difference in AUROC ($\Delta$AUROC) between full and reduced models.

The DeLong test accounts for the covariance structure of paired predictions, enabling statistically principled comparison of correlated ROC curves.  A feature set was deemed to have independent predictive value when the 95\,\% CI for $\Delta$AUROC excluded zero, \rev{that is, when both confidence bounds lay strictly on the same side of zero. $\Delta$AUROC and its confidence bounds are reported to three decimal places. Where a bound falls within 0.001 of zero the result is reported as marginal and marked as such in} Table~\ref{tab:delong_comparative_analysis}; statistically significant drops in AUROC following the removal of a feature set are interpreted as evidence for its contribution to predictive performance.

\subsection{Association Between Feedback Quality and Outcomes}
\label{sec:association_testing}

To assess how distinct feedback quality dimensions relate to subsequent \emph{trainee \revT{behavior} adjustments} or \emph{trainee verbal acknowledgment}, we fit a generalized linear mixed model (GLMM) with a Poisson distribution and log link function. The binary outcome indicated whether a feedback instance was followed by a \revT{behavioral adjustment} from the trainee. Six quality dimensions were entered as fixed effects: \revT{Encouragement, Urgency, Actionability, Timeliness, Clarity, and Reflection}. A random intercept for surgical case was included to account for clustering across repeated observations within the same case. We report exponentiated fixed effects as rate ratios, with 95\% confidence intervals. \rev{We additionally computed estimated marginal means at each point of the 1--5 rating scale for every dimension, which show the shape of each relationship rather than a single multiplicative summary; these are reported in Appendix~\ref{apx:marginal_effects}.}

\subsection{Human-AI Alignment Evaluation}
\label{sec:human_alignment}

To evaluate alignment with human judgment, we took the AI-discovered quality rating definitions and asked two human raters with domain knowledge to apply them to a stratified sample of 30 feedback instances. These instances were selected to ensure coverage across the full scoring range (1–5) for each quality dimension. Specifically, for each of the six feedback quality dimensions, we sampled five examples: two with high scores (4 or 5), one with a mid-range score (3), and two with low scores (1 or 2). To avoid repetition, already sampled examples were excluded in subsequent selections using index tracking. This process ensured that the evaluation set reflected the diversity of scoring scenarios across all criteria.

Prior to annotation, raters participated in a calibration session using a separate set of 30 feedback examples. Human-human disagreements from this session were used for discussion among annotators towards reaching interpretation consensus following best practices in qualitative coding \citep{campbell2013coding, chinh2019ways}, while human-AI disagreements were used as examples incorporated into the scoring definitions for AI behavior steering following in-context-learning principle \cite{dong2024survey}, as described in Section \Sref{sec:method_scoring_hum_ai_align}. Final inter-rater agreement results were computed using the unseen 30-instance evaluation set.

We calculated quadratically weighted Kappa score following best practices from \cite{holland2022reliability, watkins2017evaluation}. Quadratically‑weighted $\kappa$ was chosen because Behaviorally Anchored Rating Scales (BARS) deliberately place the extreme anchors much farther apart—conceptually and clinically—than adjacent mid‑points.  Quadratic weights reflect this by squaring the category distance, heavily penalizing large misclassifications, whereas linear weights down‑weight all disagreements in a strictly proportional (less discriminating) fashion \citep{viera2005understanding}. Quadratic weighting is therefore recommended for ordered clinical/BARS‑style rubrics and is the form originally proposed by \citet{cohen1968weighted}.

\backmatter

% \bmhead{Supplementary information}

% If your article has accompanying supplementary file/s please state so here. 

% Authors reporting data from electrophoretic gels and blots should supply the full unprocessed scans for key as part of their Supplementary information. This may be requested by the editorial team/s if it is missing.

% Please refer to Journal-level guidance for any specific requirements.

\section*{Data Availability}
The datasets generated during and/or analyzed during the current study are available from the corresponding author on reasonable request.

\section*{Code Availability}
All GPT-4o interactions were run using standard OpenAI API \citep{openai_api}. The precise wording of the system and user prompts at each framework stage is provided in Supplementary Materials. Agglomerative clustering analysis has been performed using a standard scikit-learn implementation \citep{Agglomer75:online}. The integration code is available from the corresponding author upon reasonable request.
Feedback effectiveness prediction analysis has been performed using standard Scikit-learn implementations including RandomForestClassifier \citep{RandomFo48:online}, GridSearchCV \citep{GridSear54:online}, and StratifiedKFold \cite{Stratifi28:online}. Inter-rater reliability was assessed using standard weighted Kappa implementation from Scikit-learn \citep{cohenkap50:online}.
Association analysis was performed using R implementations of the generalized linear mixed‐effects (\texttt{glmer}) from the \texttt{lme4} package (v1.1.37) and estimated marginal means were computed with the \texttt{emmeans} package (v1.11.2); we did not build custom code for machine learning evaluation or association analysis.

\section*{Acknowledgments}
This study was supported in part by the National Cancer Institute under Award Numbers R01CA251579 and R01CA298988. The funder had no role in the design and conduct of the study; collection, management, analysis, and interpretation of the data; preparation, review, or approval of the manuscript; and decision to submit the manuscript for publication.

\section*{Author Contributions}
R.K.: conceptualization, methodology, evaluation, experimental analysis and visualization, writing original draft, review and editing. J.E.K.: conceptualization, evaluation, review and editing
S.Y.C.: methodology, experimental analysis, writing review and editing.
J.L., C.Y.: data curation and evaluation.
A.D., U.P.: supervision, writing review and editing. P.W., A.A., J.H.: conceptualization, funding acquisition, supervision, writing review and editing.

\section*{Competing Interests}
Rafal Kocielnik declares no competing financial or non-financial interests \\
J. Everett Knudsen declares no competing financial or non-financial interests \\
Steven Y. Cen declares no competing financial or non-financial interests \\
Jasmine Lin declares no competing financial or non-financial interests \\
Cherine H. Yang declares no competing financial or non-financial interests \\
Atharva Deo declares no competing financial or non-financial interests \\
Ujjwal Pasupulety declares no competing financial or non-financial interests \\
Peter Wager declares no competing financial or non-financial interests \\
Anima Anandkumar declares no competing financial or non-financial interests \\
Andrew J. Hung declares no competing non-financial interests, but
reports financial disclosures with Intuitive Surgical, Inc. and Teleflex, Inc.

% \section*{Declarations}
% Some journals require declarations to be submitted in a standardised format. Please check the Instructions for Authors of the journal to which you are submitting to see if you need to complete this section. If yes, your manuscript must contain the following sections under the heading `Declarations':

% \begin{itemize}
% \item Conflict of interest/Competing interests (check journal-specific guidelines for which heading to use)
% \item Ethics approval and consent to participate
% \item Consent for publication
% \item Materials availability
% \end{itemize}

%\noindent
%If any of the sections are not relevant to your manuscript, please include the heading and write `Not applicable' for that section. 

% \bigskip

% Springer journals and proceedings: \url{https://www.springer.com/gp/editorial-policies}

% Nature Portfolio journals: \url{https://www.nature.com/nature-research/editorial-policies}

% \textit{Scientific Reports}: \url{https://www.nature.com/srep/journal-policies/editorial-policies}

% BMC journals: \url{https://www.biomedcentral.com/getpublished/editorial-policies}
% \end{flushleft}

\begin{appendices}

%%%%%%%%%%%%%%%%%%%%%%%%%%%%%%%%%%%%%%%%%%%%%%%%%%%%%%%%%%%%%%%%%%%%%%%%
%% Make hyperref anchors unique inside the appendices.
%%
%% The appendix float counters restart (so tables print as A1, C1, E1 and
%% figures as D1, E1), but hyperref builds its anchor names from the RAW
%% counter value, not from \thetable/\thefigure. Without this, appendix
%% Table C1 and body Table 1 both claim the anchor "table.1", and every
%% \ref to an appendix float silently jumps to the body float instead.
%% Keying \theHtable/\theHfigure on the section letter makes them unique
%% (table.C.1, figure.D.1, ...). Printed numbering is unaffected.
%%%%%%%%%%%%%%%%%%%%%%%%%%%%%%%%%%%%%%%%%%%%%%%%%%%%%%%%%%%%%%%%%%%%%%%%
\renewcommand{\theHtable}{\thesection.\arabic{table}}
\renewcommand{\theHfigure}{\thesection.\arabic{figure}}

\section{Discovered Feedback Quality Scoring Criteria}
\label{apx:discovered_criteria}

%% Discovered Criteria Definitions p1 %%
\begin{table}[h!]
\centering
\small
\begin{tabular}{p{1.4cm}p{3.3cm}p{3.3cm}p{3.3cm}}
\toprule
\textbf{Scale} & \textbf{Score 1 (None)} & \textbf{Score 3 (Moderate)} & \textbf{Score 5 (High)} \\
\midrule

\rowcolor{gray!20}
\multicolumn{4}{l}{\textbf{Encouragement}} \\

Definition & 
\raggedright Neutral or negative tone; no observable encouragement. &
\raggedright Mild affirmations or neutral support without strong reinforcement. &
\raggedright Explicit, enthusiastic encouragement clearly aimed at boosting confidence. \\

\addlinespace[4pt]

Examples &
\parbox[t]{\linewidth}{\raggedright 
\textit{“That’s wrong.”} \\
\textit{“Yes, it’s to define the bladder neck…”} \\
\textit{“Do you see where the pelvic bone is now?”}
} &
\parbox[t]{\linewidth}{\raggedright 
\textit{“That’s fine.”} \\
\textit{“Good.”} \\
\textit{“Keep going, smiley face”}
} &
\parbox[t]{\linewidth}{\raggedright 
\textit{“Excellent work!”} \\
\textit{“Perfect.”} \\
\textit{“Good, I love that.”}
} \\

\midrule
\rowcolor{gray!20}
\multicolumn{4}{l}{\textbf{Urgency}} \\

Definition & 
\raggedright Calm or delayed feedback; no urgency communicated. &
\raggedright Prompting language or tone suggests some urgency without direct commands. &
\raggedright Explicit, critical urgency with immediate calls to action. \\

\addlinespace[4pt]

Examples &
\parbox[t]{\linewidth}{\raggedright 
\textit{“You might want to adjust that later.”} \\
\textit{“This one should have been further distal.”} \\
\textit{“Maybe next time.”}
} &
\parbox[t]{\linewidth}{\raggedright 
\textit{“Watch your positioning.”} \\
\textit{“You go a little more.”} \\
\textit{“Let’s work distally.”}
} &
\parbox[t]{\linewidth}{\raggedright 
\textit{“Immediately stop!”} \\
\textit{“Correct your grip now!”} \\
\textit{“Don’t coag!!”}
} \\

\midrule
\rowcolor{gray!20}
\multicolumn{4}{l}{\textbf{Actionability}} \\

Definition & 
\raggedright No actionable content; vague or evaluative. &
\raggedright Some guidance, but lacks precision. &
\raggedright Highly precise, step-by-step action that is immediately executable. \\

\addlinespace[4pt]

Examples &
\parbox[t]{\linewidth}{\raggedright 
\textit{“That’s not right.”} \\
\textit{“This one should have been even further distal.”} \\
\textit{“Be more careful.”}
} &
\parbox[t]{\linewidth}{\raggedright 
\textit{“Adjust your grip.”} \\
\textit{“Keep going, smiley face.”} \\
\textit{“Ok, open that up.”}
} &
\parbox[t]{\linewidth}{\raggedright 
\textit{“Move your needle holder 2 cm forward.”} \\
\textit{“Angle it down by 30 degrees.”} \\
\textit{“Sweep left and then buzz.”}
} \\

\bottomrule
\end{tabular}
\caption{AI-discovered feedback quality scoring criteria. Each dimension is rated using a 5-point Behaviorally Anchored Rating Scale (BARS); for clarity, only anchors for scores 1 (Minimal), 3 (Moderate), and 5 (High) are shown. Definitions and representative examples are provided for each anchor.}

\label{tab:rubric_feedback_p1}
\end{table}

%%% Continuation of the Discovered Criteria Table

\begin{table}[ht!]
\centering
\small
\begin{tabular}{p{1.4cm}p{3.3cm}p{3.3cm}p{3.3cm}}
\toprule
\textbf{Scale} & \textbf{Score 1 (None)} & \textbf{Score 3 (Moderate)} & \textbf{Score 5 (High)} \\

\midrule
\rowcolor{gray!20}
\multicolumn{4}{l}{\textbf{Timeliness}} \\

Definition & 
\raggedright Feedback is delayed significantly, unrelated to immediate action. &
\raggedright Moderately prompt; refers to recent action but not immediate. &
\raggedright Immediate, real-time feedback during the action. \\

\addlinespace[4pt]

Examples &
\parbox[t]{\linewidth}{\raggedright 
\textit{“This one should have been even further distal.”} \\
\textit{“Maybe next time.”} \\
\textit{“Before you continue, let’s review.”}
} &
\parbox[t]{\linewidth}{\raggedright 
\textit{“So you start here.”} \\
\textit{“This is where I want you to start.”} \\
\textit{“You did that too fast.”}
} &
\parbox[t]{\linewidth}{\raggedright 
\textit{“It’s still bleeding—you gotta stop that.”} \\
\textit{“Do you see exactly where the pelvic bone is now?”} \\
\textit{“You’re coag’ing again!”}
} \\

\midrule

\rowcolor{gray!20}
\multicolumn{4}{l}{\textbf{Clarity}} \\

Definition & 
\raggedright Confusing or ambiguous; unclear what is meant. &
\raggedright Mostly clear but includes minor ambiguities. &
\raggedright Exceptionally clear, concise, and unambiguous. \\

\addlinespace[4pt]

Examples &
\parbox[t]{\linewidth}{\raggedright 
\textit{“You know what to do.”} \\
\textit{“Good.”} \\
\textit{“Let me see how much is bleeding.”}
} &
\parbox[t]{\linewidth}{\raggedright 
\textit{“Adjust it a bit.”} \\
\textit{“That’s it, that’s all you’re gonna get.”} \\
\textit{“Ok, open that up.”}
} &
\parbox[t]{\linewidth}{\raggedright 
\textit{“Insert the needle at a 45-degree angle just above the marked point.”} \\
\textit{“How many knots did you do there?”} \\
\textit{“Move the instrument to the left.”}
} \\

\midrule
\rowcolor{gray!20}
\multicolumn{4}{l}{\textbf{Reflection}} \\

Definition & 
\raggedright No reflective element; purely directive or evaluative. &
\raggedright Occasional reflective prompt but lacks depth. &
\raggedright Strong, open-ended reflective guidance fostering deep evaluation. \\

\addlinespace[4pt]

Examples &
\parbox[t]{\linewidth}{\raggedright 
\textit{“Before you do the next step, clean your lens.”} \\
\textit{“Closer to the prostate.”} \\
\textit{“Push a little further.”}
} &
\parbox[t]{\linewidth}{\raggedright 
\textit{“What went wrong there?”} \\
\textit{“Do you see that periurethral stuff?”} \\
\textit{“Do you see where the pelvic bone is now?”}
} &
\parbox[t]{\linewidth}{\raggedright 
\textit{“How could you adjust your technique to improve precision?”} \\
\textit{“What’s the next step?”} \\
\textit{“Why did you choose that approach?”}
} \\

\bottomrule
\end{tabular}
\caption{Scoring criteria for the final two dimensions, \textbf{Clarity} and \textbf{Reflection}, discovered via AI-based analysis of surgical feedback. Anchors correspond to scores 1 (Minimal), 3 (Moderate), and 5 (High) on a 5-point Behaviorally Anchored Rating Scale (BARS).}
\label{tab:rubric_feedback_p2}
\end{table}

\newpage

\section{GPT-4o prompts}
\label{apx:llm_prompts}

%%%%%%%
% PROMPT 1 - Criteria Discovery
%%%%%%%

\begin{table}[h!]
  \centering
  \caption{\textbf{Prompt used to elicit new quality-rating dimensions for surgical feedback.}}
  \vspace{-1em}
  \label{tab:quality_criteria_discovery_prompt}
  \begin{tcolorbox}[
    sharp corners,
    colback=white,
    colframe=black!75!white,
    coltitle=white,
    fonttitle=\bfseries,
    boxrule=0.5pt,
    title=Prompt Template for Quality-Criteria Discovery,
    width=\textwidth
  ]
\textbf{System Instruction:}\\
You are working in the context of \emph{verbal feedback} delivered by a trainer to a trainee in a live surgery.  
The goal of the feedback is to modify trainee thinking or behavior.  
There are different measures assessing feedback effectiveness, including:  
\begin{itemize}
  \item \textbf{Trainee Behavior Change} — behavioral adjustment made by the trainee that corresponds directly with the preceding feedback (e.g.\ trainee immediately pulls more tightly on the suture thread after receiving feedback to cinch tightly);
  \item \textbf{Trainee Verbal Acknowledgment} — verbal or audible confirmation by the trainee confirming that they have heard the feedback (e.g.\ “Okay, I see”, “uh-huh, got it.”);
  \item \textbf{Trainer Approval} — trainer verbally demonstrates that they are satisfied with the trainee behavioral change (e.g.\ “yes”, “mhm”).
\end{itemize}

Based on these descriptions, propose dimensions that would be predictive of the three outcomes above.  
For each dimension, supply a definition such that a rater could score a feedback instance on the Behaviorally Anchored Rating Scale (BARS) using 5 behavioral anchor levels, from  
\textbf{1 = feedback does \emph{not} exhibit this quality} to  
\textbf{5 = feedback clearly possesses this quality}.  
Dimensions must be applicable to \emph{transcribed feedback lines alone}—without preceding dialogue, video, or timing information.

\medskip
\textbf{User Message:}\\
Produce an output in the format:\\
\texttt{No|Dimension Name|Scoring Definition|Score 1 rating|Score 3 rating|Score 5 rating}\\
Do \textbf{not} include this header in your reply.
  \end{tcolorbox}
\end{table}

%%%%%%%
% PROMPT 2 - Criterie Consolidaiton Prompt
%%%%%%%

\begin{table}[ht]
  \centering
  \caption{\textbf{Prompt used to merge overlapping quality‑rating criteria into a single consolidated dimension.}}
  \vspace{-1em}
  \label{tab:criteriaConsolidationPrompt}
  \begin{tcolorbox}[
    sharp corners,
    colback=white,
    colframe=black!75!white,
    coltitle=white,
    fonttitle=\bfseries,
    boxrule=0.5pt,
    title=Prompt Template for Criteria Consolidation Phrasing per Cluster,
    width=\textwidth
  ]
\textbf{System Instruction:}\\
You are given a set of \emph{similar scoring criteria}, each with a name and definition.  
Combine them under one unified name and definition.  
Consolidate into \textbf{exactly one} refined criterion based on the list below:\\

Name: \texttt{[Criterion 1 Name]}, Definition: \texttt{[Criterion 1 Definition]}\\
Name: \texttt{[Criterion 2 Name]}, Definition: \texttt{[Criterion 2 Definition]}\\
\ldots \\[0.5em]

\textbf{User Message:}\\
Based on the consolidated and refined criterion, output a Python tuple in the form:\\
\texttt{(No, "Consolidated Name", "Consolidated Definition")}\\
Only return the tuple—no additional commentary.
  \end{tcolorbox}
\end{table}

%%%%%%%
% PROMPT 3 - Feedback Scoring - LLM as a judge
%%%%%%%
\begin{table}[ht]
  \centering
  \caption{\textbf{Prompt used for scoring surgical feedback on multiple quality criteria.}}
  \vspace{-1em}
  \label{tab:surgicalFeedbackScoringPrompt}
  \begin{tcolorbox}[
    sharp corners,
    colback=white,
    colframe=black!75!white,
    coltitle=white,
    fonttitle=\bfseries,
    boxrule=0.5pt,
    title=Prompt Template for Multi-Criteria Feedback Scoring,
    width=\textwidth
  ]
\textbf{System Instruction:}\\
This is verbal FEEDBACK delivered during surgery by a trainer to a trainee.\\
Please rate it given each of the following criteria and associated scales.\\

Q1. [Criterion 1]\\
Q2. [Criterion 2]\\
Q3. [Criterion 3]\\
Q4. [Criterion 4]\\
Q5. [Criterion 5]\\
Q6. [Criterion 6]\\

Make the scoring concise as it needs to be parsed automatically later on; use the format of an ordered Python list, don't repeat question numbers:\\
\textit{Q1 score, Q2 score, Q3 score, Q4 score, Q5 score, Q6 score}
%e.g., \texttt{[1, 2, 1, 2, 1]}\\

\medskip
\textbf{User Message:}\\
FEEDBACK: ``[feedback\_line]''
  \end{tcolorbox}
\end{table}

%%%%%%%%%%%%%%%%%%%%%%%%%%%%%%%%%%%%%%%%%%%%%%%%%%%%%%%%%%%%%%%%%%%%%%%%
%% NEW APPENDIX (R2): correlation structure, multicollinearity and
%% cross-model (open-weight) replication.
%%%%%%%%%%%%%%%%%%%%%%%%%%%%%%%%%%%%%%%%%%%%%%%%%%%%%%%%%%%%%%%%%%%%%%%%
\newpage
\section{\rev{Correlation Structure Among Quality Criteria}}
\label{apx:correlation_structure}
\setcounter{table}{0}% number this appendix's tables C1, C2, C3

\rev{This appendix reports the correlation structure among the six finalized rubric scores, the associated multicollinearity diagnostics, the sensitivity of predictive performance to surgical-case grouping, and the agreement obtained when the finalized rubric is applied by two open-weight models.}

%% Table C1 - Spearman correlations
\begin{table}[h!]
\centering
\small
\caption{\rev{Spearman correlations among the six quality criteria scored by GPT-4o across 4,210 feedback instances. Column abbreviations follow the row order.}}
\label{tab:spearman_criteria}
% numbers right-aligned with a phantom minus on non-negative values, so the
% decimal points line up down every column
\newcommand{\pz}{\phantom{$-$}}
\begin{tabular}{l*{6}{r@{\hskip 0.28in}}}
\toprule
\textbf{Dimension} & \multicolumn{1}{c@{\hskip 0.28in}}{\textbf{Enc}} & \multicolumn{1}{c@{\hskip 0.28in}}{\textbf{Urg}} & \multicolumn{1}{c@{\hskip 0.28in}}{\textbf{Act}} & \multicolumn{1}{c@{\hskip 0.28in}}{\textbf{Tim}} & \multicolumn{1}{c@{\hskip 0.28in}}{\textbf{Cla}} & \multicolumn{1}{c@{\hskip 0.28in}}{\textbf{Refl}} \\
\midrule
Encouragement & \pz--    & \pz0.17 & \pz0.41 & $-0.16$ & \pz0.30 & $-0.05$ \\
Urgency       & \pz0.17  & \pz--   & \pz0.63 & \pz0.33 & \pz0.54 & \pz0.02 \\
Actionability & \pz0.41  & \pz0.63 & \pz--   & \pz0.18 & \pz0.61 & $-0.01$ \\
Timeliness    & $-0.16$  & \pz0.33 & \pz0.18 & \pz--   & \pz0.27 & $-0.12$ \\
Clarity       & \pz0.30  & \pz0.54 & \pz0.61 & \pz0.27 & \pz--   & \pz0.13 \\
Reflection    & $-0.05$  & \pz0.02 & $-0.01$ & $-0.12$ & \pz0.13 & \pz--   \\
\bottomrule
\end{tabular}
\end{table}

%% Table C2 - VIF and case-grouping sensitivity
\begin{table}[h!]
\centering
\small
\caption{\rev{Multicollinearity and sensitivity to surgical-case grouping for GPT-4o, Mistral Small 3.2 (24B), and Qwen3-Next 80B-A3B using the finalized rubric. VIF = variance inflation factor; the maximum is taken across the six criteria. $\Delta$AUROC = case-grouped minus outcome-stratified area under the receiver operating characteristic curve, calculated before rounding. Both schemes used five folds and a fixed 500-tree Random Forest with the six scores as predictors; grouped folds kept all utterances from each of 32 surgical cases together.}}
\label{tab:vif_grouping}
\begin{tabular}{lccc}
\toprule
\textbf{Model} & \textbf{Max. VIF} & \shortstack{\textbf{$\Delta$AUROC} \\ \textbf{Behavioral Adj.}} & \shortstack{\textbf{$\Delta$AUROC} \\ \textbf{Verbal Ack.}} \\
\midrule
GPT-4o 2024-08-06     & 2.25 & $-0.009$ & $+0.002$ \\
Mistral Small 3.2 24B & 3.29 & $-0.005$ & $+0.000$ \\
Qwen3-Next 80B-A3B    & 3.95 & $-0.007$ & $-0.012$ \\
\bottomrule
\end{tabular}
\end{table}

%% Table C3 - cross-model performance and agreement
\begin{table}[h!]
\centering
\small
\caption{\rev{Predictive performance and cross-model agreement for GPT-4o, Mistral Small 3.2 (24B), and Qwen3-Next 80B-A3B. \textbf{(A)} AUROC (95\% CI), using the setup and sample sizes in Table~\ref{tab:vif_grouping}; CIs use 1,000 percentile-bootstrap resamples of individual instances from pooled out-of-fold predictions for both splitting schemes. \textbf{(B)} Agreement with the GPT-4o rerun: quadratic-weighted $\kappa$ (95\% CI) for each criterion and their arithmetic mean, followed by the percentage of scores within one point on the 1--5 scale. Items were matched by ID ($n = 4{,}210$), spanning 32 surgical cases. Agreement CIs use 2,000 percentile-bootstrap resamples of cases, retaining all utterances and all six scores together; mean $\kappa$ was recalculated in each resample.}}
\label{tab:cross_model}
\vspace{0.3em}

\rev{\textbf{A. Predictive performance}}

\vspace{0.3em}
{\footnotesize
\begin{tabular}{@{}p{2.5cm}p{2.9cm}cc@{}}
\toprule
\textbf{Outcome} & \textbf{Scoring model} & \shortstack{\textbf{Stratified AUROC} \\ \textbf{(95\% CI)}} & \shortstack{\textbf{Case-grouped AUROC} \\ \textbf{(95\% CI)}} \\
\midrule
\multirow{3}{2.6cm}{Behavioral Adjustment}
  & GPT-4o             & 0.741 (0.725--0.756) & 0.732 (0.717--0.747) \\
  & Mistral Small 3.2  & 0.727 (0.713--0.742) & 0.722 (0.708--0.738) \\
  & Qwen3-Next 80B-A3B & 0.731 (0.716--0.746) & 0.723 (0.709--0.738) \\
\midrule
\multirow{3}{2.6cm}{Verbal Acknowledgment}
  & GPT-4o             & 0.692 (0.675--0.708) & 0.694 (0.678--0.711) \\
  & Mistral Small 3.2  & 0.699 (0.683--0.715) & 0.699 (0.684--0.716) \\
  & Qwen3-Next 80B-A3B & 0.675 (0.658--0.690) & 0.662 (0.645--0.677) \\
\bottomrule
\end{tabular}
}

\vspace{1em}
\rev{\textbf{B. Agreement with the GPT-4o rerun}}

\vspace{0.3em}
\begin{tabular}{lcc}
\toprule
\textbf{Measure} & \textbf{Mistral Small 3.2} & \textbf{Qwen3-Next 80B-A3B} \\
\midrule
Encouragement & 0.254 (0.230--0.281) & 0.552 (0.527--0.578) \\
Urgency       & 0.716 (0.692--0.738) & 0.706 (0.680--0.730) \\
Actionability & 0.668 (0.648--0.688) & 0.783 (0.770--0.799) \\
Timeliness    & 0.379 (0.338--0.418) & 0.263 (0.219--0.309) \\
Clarity       & 0.652 (0.631--0.670) & 0.630 (0.605--0.654) \\
Reflection    & 0.500 (0.459--0.539) & 0.664 (0.625--0.696) \\
\midrule
\textbf{Mean $\kappa$ across six criteria} & \textbf{0.528 (0.514--0.541)} & \textbf{0.600 (0.587--0.612)} \\
Scores within one point & 89.1\% (88.0\%--90.1\%) & 90.5\% (89.7\%--91.2\%) \\
\bottomrule
\end{tabular}
\end{table}

%%%%%%%%%%%%%%%%%%%%%%%%%%%%%%%%%%%%%%%%%%%%%%%%%%%%%%%%%%%%%%%%%%%%%%%%
%% NEW APPENDIX: per-score marginal effects behind the rate ratios
%% reported in Figure 2.
%%%%%%%%%%%%%%%%%%%%%%%%%%%%%%%%%%%%%%%%%%%%%%%%%%%%%%%%%%%%%%%%%%%%%%%%
\clearpage % flush Appendix C's floats first, so this heading and its hyperref
           % anchor land on the same page (\newpage alone leaves the anchor behind)
\section{\rev{Marginal Effects of Feedback Quality Scores}}
\label{apx:marginal_effects}

\begin{figure}[h!]
    \centering
    \includegraphics[width=\linewidth]{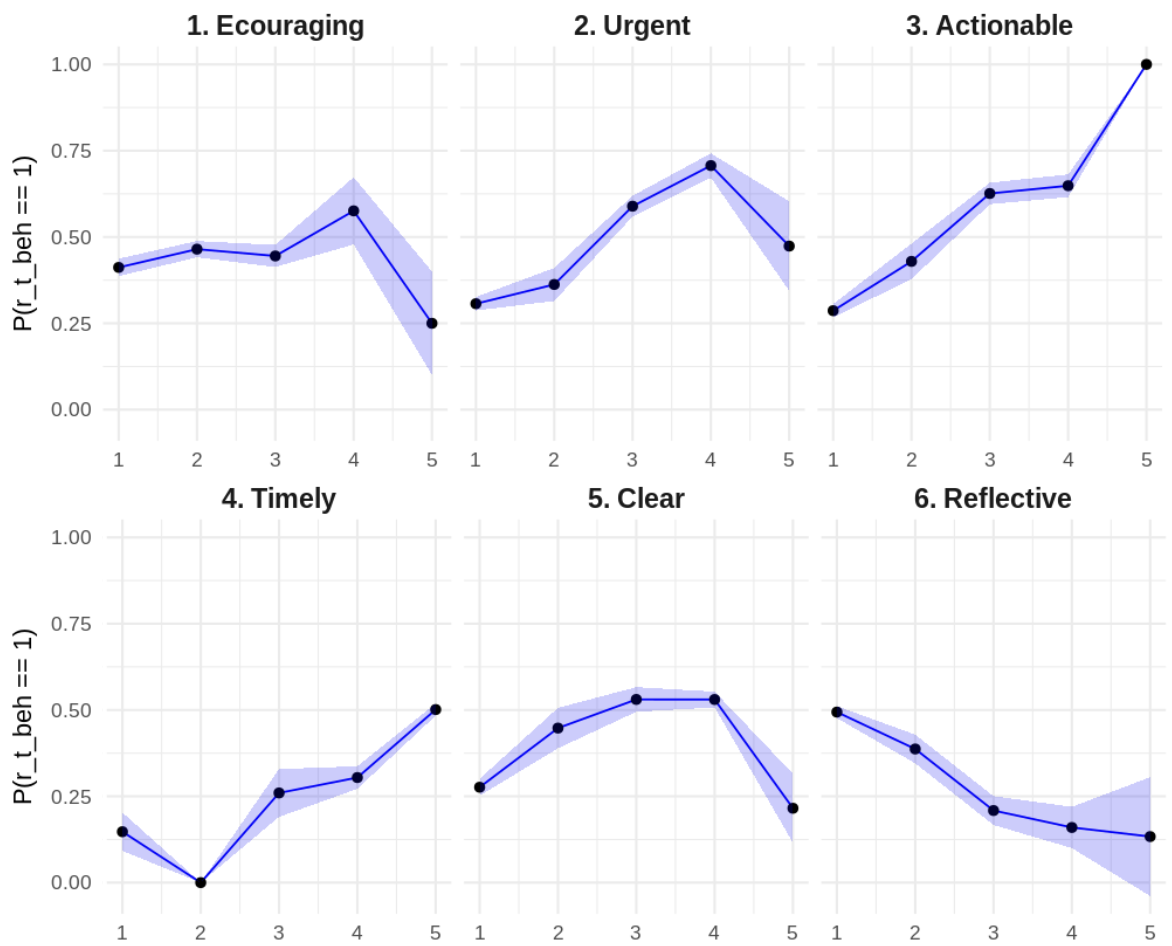}
    \caption{\rev{\textbf{Estimated marginal probability of \emph{Trainee Behavioral Adjustment} by feedback quality dimension and BARS score.} Each panel varies one of the six quality dimensions across its 1--5 Behaviorally Anchored Rating Scale while holding the others at their observed distribution; the horizontal axis of every panel is that 1--5 scale. Shaded bands are 95\% confidence intervals. The monotone rise for \emph{Actionability} corresponds to the largest positive rate ratio in Figure~\ref{fig:quality_associations}a, and the monotone decline for \emph{Reflection} to the largest negative one. Wide intervals at the extreme anchors reflect the small number of feedback instances scored there.}}
    \label{fig:marginal_effects}
\end{figure}

\begin{figure}[h!]
    \centering
    \includegraphics[width=\linewidth]{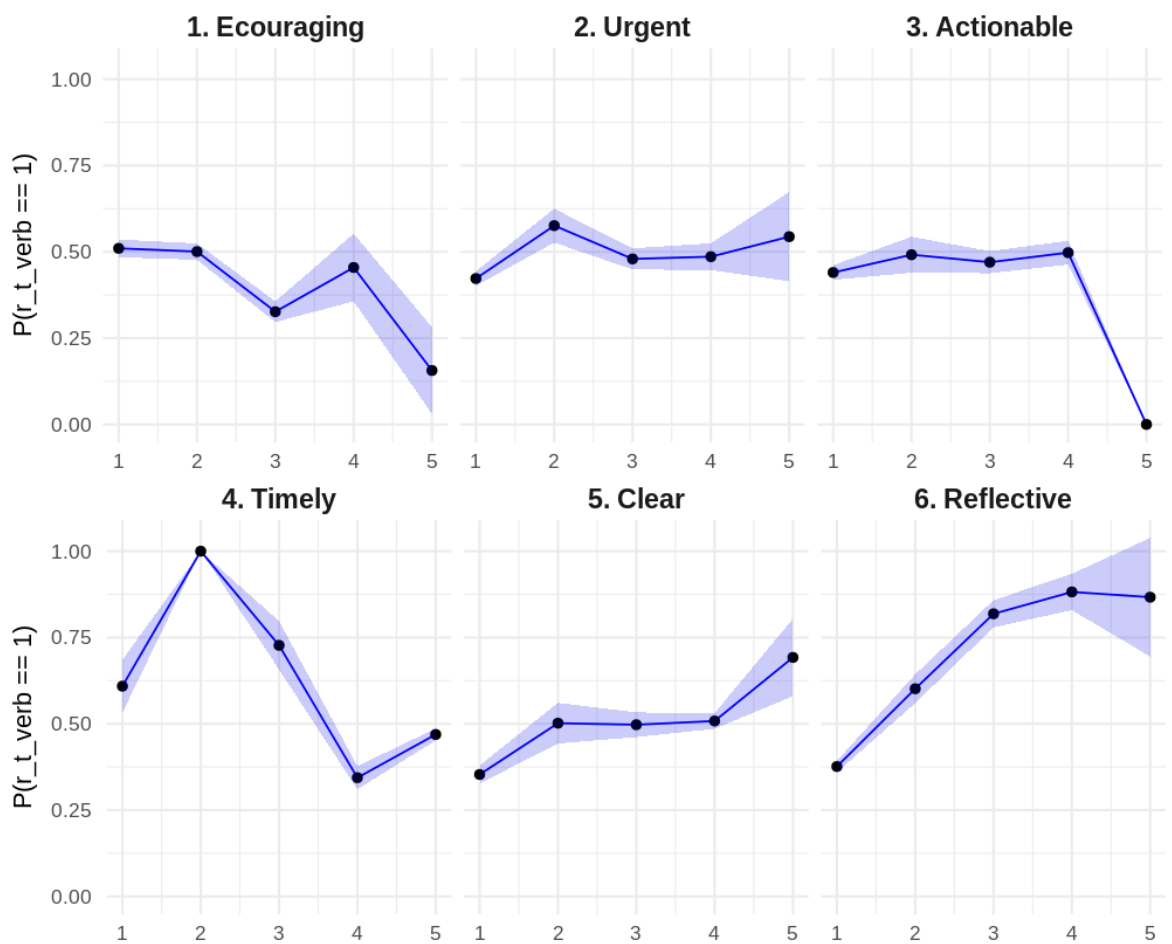}
    \caption{\rev{\textbf{Estimated marginal probability of \emph{Trainee Verbal Acknowledgment} by feedback quality dimension and BARS score.} Axes, bands and conditioning are as in Figure~\ref{fig:marginal_effects}. The monotone rise for \emph{Reflection} corresponds to the largest positive rate ratio in Figure~\ref{fig:quality_associations}b, while \emph{Encouragement} declines across the scale.}}
    \label{fig:marginal_effects_verbal}
\end{figure}

\rev{The rate ratios in Figure~\ref{fig:quality_associations} summarize each quality dimension with a single multiplicative effect, which assumes the effect is monotone across the rating scale. Figures~\ref{fig:marginal_effects} and~\ref{fig:marginal_effects_verbal} show the underlying estimated marginal means at each BARS score, so the shape of each relationship can be inspected directly. Estimates are from the same Poisson GLMM described in \Sref{sec:association_testing}, with a random intercept for surgical case; marginal means were computed with the \texttt{emmeans} package. Bands are 95\% confidence intervals, which widen at scale points that are rarely used.}

% Shown upright as two figures, three panels per row. The original single
% 12-panel image is too wide to be legible at text width, and rotating it
% made it harder to read rather than easier.

\clearpage % flush appendix floats so they cannot drift into the reference list

%%%%%%%%%%%%%%%%%%%%%%%%%%%%%%%%%%%%%%%%%%%%%%%%%%%%%%%%%%%%%%%%%%%%%%%%
%% NEW APPENDIX: the clustering behind the consolidation step (Fig. 1 b2)
%%%%%%%%%%%%%%%%%%%%%%%%%%%%%%%%%%%%%%%%%%%%%%%%%%%%%%%%%%%%%%%%%%%%%%%%
\section{\rev{Clustering Behind Criteria Consolidation}}
\label{apx:consolidation_clustering}
\setcounter{figure}{0}% number this appendix's figure E1, not continuing from D
\setcounter{table}{0}% likewise for its table

\begin{figure}[h!]
    \centering
    \includegraphics[width=0.82\linewidth]{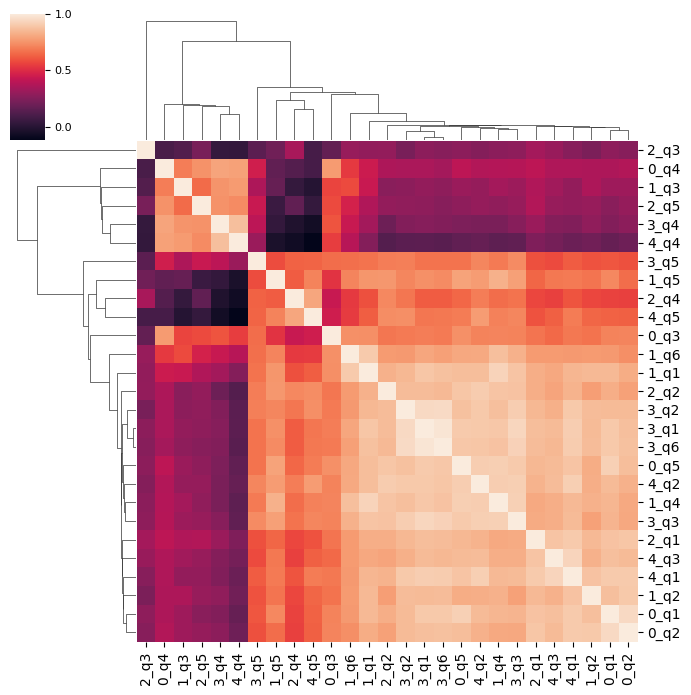}
    \caption{\rev{\textbf{Correlation structure and hierarchical clustering of the candidate criteria proposed across the five instances.} Cells give the correlation between the scores two candidate criteria assign to the same 4{,}210 feedback instances, so criteria that behave alike are adjacent regardless of their wording. Row and column dendrograms use single-linkage agglomerative clustering with Euclidean distance over the correlation profiles. Labels are \texttt{<instance>\_q<criterion>}. The large light block is the set of broadly intercorrelated candidates, while the darker off-diagonal blocks mark candidate criteria that score feedback in distinctly different ways.}}
    \label{fig:criteria_clustermap}
\end{figure}

\rev{The consolidation stage (Figure~\ref{fig:task_overview}, b2) groups the candidate criteria proposed by the five independent instances according to how similarly they score the same feedback, rather than by how similarly they are worded. Figure~\ref{fig:criteria_clustermap} shows the matrix and dendrogram that step operates on. Each row and column is one candidate criterion, labelled \texttt{<instance>\_q<criterion>}, where the leading index identifies which of the five instances proposed it and the \texttt{q} index numbers that instance's own proposed dimensions. Cutting the dendrogram at $k=6$, the value selected by the silhouette criterion described in \Sref{sec:ai_discovery}, yields the six groups that were each unified into one of the final rubric dimensions. Table~\ref{tab:candidate_clusters} lists the candidate criteria in each cluster under the names their proposing instance gave them, showing which qualities were proposed repeatedly and which emerged only once.}

%% Candidate criteria -> cluster mapping (rows generated from crit_df)
\begin{table}[h!]
\centering
\footnotesize
\caption{\rev{\textbf{Candidate criteria proposed by each instance, grouped by consolidation cluster.} Each entry gives the name an instance assigned to its own candidate criterion, followed by the identifiers used in Figure~\ref{fig:criteria_clustermap}. \emph{Instances} counts how many of the five proposed a criterion that fell into the cluster. Because clustering is on how criteria score feedback rather than on their names, a name may recur across instances without its instances falling in one cluster: \emph{Clarity} was proposed by all five, but four of those candidates score like the actionability group and only one forms a cluster of its own, and the two \emph{Promptness} candidates separate into different clusters.}}
\label{tab:candidate_clusters}
\begin{tabular}{@{}lccp{6.3cm}@{}}
\toprule
\textbf{Cluster} & \textbf{Candidates} & \textbf{Instances} & \textbf{Criterion name as proposed (identifier)} \\
\midrule
1 --- \emph{Encouragement} & 5 & 5/5 & Encouragement/Supportiveness (0\_q4); Encouragement (1\_q3, 3\_q4, 4\_q4); Supportiveness (2\_q5) \\
\addlinespace[2pt]
2 --- \emph{Urgency} & 3 & 3/5 & Urgency (1\_q5); Promptness (2\_q4); Urgency/Immediacy (4\_q5) \\
\addlinespace[2pt]
3 --- \emph{Actionability} & 16 & 5/5 & Specificity of Feedback (0\_q1); Actionability of Feedback (0\_q2); Relevance to the Task (0\_q5); Clarity of Instruction (1\_q1, 2\_q1, 4\_q1); Actionability (1\_q2, 3\_q2); Relevance (1\_q4, 3\_q3); Trainer Engagement (1\_q6); Relevance to Task (2\_q2, 4\_q2); Clarity of Feedback (3\_q1); Specificity (3\_q6); Constructiveness (4\_q3) \\
\addlinespace[2pt]
4 --- \emph{Clarity} & 1 & 1/5 & Clarity of Feedback (0\_q3) \\
\addlinespace[2pt]
5 --- \emph{Timeliness} & 1 & 1/5 & Promptness (3\_q5) \\
\addlinespace[2pt]
6 --- \emph{Reflection} & 1 & 1/5 & Encouragement of Self-assessment (2\_q3) \\
\bottomrule
\end{tabular}
\end{table}

\clearpage

\end{appendices}

\newpage
\bibliography{bibliography}% common bib file

\end{document}